\pdfoutput=1

\documentclass[11pt]{article}

\usepackage[final]{acl}

\usepackage{fancyhdr}
\fancypagestyle{firstpage}{
  \fancyhf{}
    \fancyhead[L]{\normalsize Accepted in ACL 2025}
}

\usepackage{times}
\usepackage{latexsym}
\usepackage{multirow} 
\usepackage{booktabs} 
\usepackage{booktabs}
\usepackage{adjustbox}
\usepackage[T1]{fontenc}

\usepackage[utf8]{inputenc}

\usepackage{microtype}

\usepackage{inconsolata}

\usepackage{graphicx}
\usepackage{subcaption}
\usepackage{float}
\usepackage{makecell} 

\title{MVTamperBench: Evaluating Robustness of Vision-Language Models}

\author{
 \textbf{Amit Agarwal\textsuperscript{1*+}},
 \textbf{Srikant Panda\textsuperscript{2*}},
 \textbf{Angeline Charles\textsuperscript{3*}},
 \textbf{Hitesh Laxmichand Patel\textsuperscript{5}},\\
 \textbf{Bhargava Kumar\textsuperscript{4}},
 \textbf{Priyaranjan Pattnayak\textsuperscript{6}},
 \textbf{Taki Hasan Rafi\textsuperscript{7}},
 \textbf{Tejaswini Kumar\textsuperscript{4}}, \\
 \textbf{Hansa Meghwani\textsuperscript{1}},
 \textbf{Karan Gupta\textsuperscript{5}},
 \textbf{Dong-Kyu Chae\textsuperscript{7+}}
\\
\\
 \textsuperscript{1}Liverpool John Moores University
 \textsuperscript{2}Birla Institute of Technology
 \textsuperscript{3}Christ University
 \\
 \textsuperscript{4}Columbia University
 \textsuperscript{5}New York University
 \textsuperscript{6}University of Washington
 \textsuperscript{7}Hanyang University
  \\
 \textsuperscript{*}Equal Contribution
 \textsuperscript{+}Corresponding Authors
\\
 \small{
   \textbf{Correspondence:} \href{mailto:email@domain}{amit.pinaki@gmail.com, dongkyu@hanyang.ac.kr}
 }
}
\usepackage{float} 
\usepackage{geometry} 
\usepackage{tabularx} 
\usepackage{array} 
\usepackage{caption} 
\newcolumntype{L}[1]{>{\raggedright\arraybackslash}p{#1}}

\begin{document}
\thispagestyle{firstpage}
\pagestyle{firstpage}
\maketitle
\begin{abstract}

Multimodal Large Language Models (MLLMs), are recent advancement of  Vision-Language Models (VLMs) that have driven major advances in video understanding. However, their vulnerability to adversarial tampering and manipulations remains underexplored. To address this gap, we introduce \textbf{MVTamperBench}, a benchmark that systematically evaluates MLLM robustness against five prevalent tampering techniques: rotation, masking, substitution, repetition, and dropping; based on real-world visual tampering scenarios such as surveillance interference, social media content edits, and misinformation injection. MVTamperBench comprises ~3.4K original videos, expanded into over ~17K tampered clips covering 19 distinct video manipulation tasks. This benchmark challenges models to detect manipulations in spatial and temporal coherence. We evaluate 45 recent MLLMs from 15+ model families. We reveal substantial variability in resilience across tampering types and show that larger parameter counts do not necessarily guarantee robustness. MVTamperBench sets a new benchmark for developing tamper-resilient MLLM in safety-critical applications, including detecting clickbait, preventing harmful content distribution, and enforcing policies on media platforms. We release all code, data, and benchmark to foster open research in trustworthy video understanding.


\end{abstract}

\section{Introduction}


Multimodal Large Language Models (MLLMs) have catalyzed significant progress in video understanding, enabling a wide array of applications across domains such as surveillance, healthcare, and autonomous systems. However, their growing integration into high-traffic platforms (e.g., Instagram, Facebook, TikTok) has exposed critical vulnerabilities. Specifically, tampered videos are increasingly exploited to bypass platform policies, disseminate harmful content, and promote clickbait, posing serious challenges for content moderation and policy enforcement \cite{kingra2023emergence,timesofindia2025youtube}.

These threats underscore the urgent need to improve the robustness of MLLMs against real-world manipulations. Although video tampering can involve audio modifications, synthetic speech overlays, or deepfake generation, our benchmark focuses exclusively on visual-only manipulations. This choice is driven by the current limitations of existing models, most of which currently lack comprehensive audiovisual processing capabilities. 

Unlike adversarial robustness in static images, an area that has been extensively studied, video tampering introduces unique challenges that arise from the interplay of spatial and temporal dynamics. Common manipulation techniques such as frame dropping, masking, repetition, substitution, and rotation disrupt this coherence, frequently resulting in catastrophic model failures. These methods mirror real-world adversarial tactics: substitution injects objectionable material (e.g., nudity, violence) to circumvent detection; dropping eliminates key surveillance evidence; repetition loops footage to obscure illicit activity; masking occludes critical regions; and rotation induces spatial distortions commonly seen in edited or re-uploaded content. 

Existing adversarial and traditional approaches, including black-box~\cite{Jiang2019BlackboxAA} and cross-modal~\cite{Wei2021CrossModalTA} attacks, primarily address isolated scenarios rather than systematically evaluating diverse tampering types. Furthermore, existing multimodal benchmarks \cite{pattnayak2024survey, agarwal2021evaluate}—such as MMBench-Video \cite{Fang:24}, BLINK \cite{Fu:25}, and Video-MME \cite{Fu:24-VideoMME}—focus on multimodal comprehension and temporal reasoning but overlook \emph{adversarial} robustness. For example, MMBench-Video evaluates cross-modal alignment without adversarial testing; BLINK targets long-form temporal reasoning without addressing tampering impacts; and Video-MME, while effective for vision-language alignment, omits tampering-specific tasks. This leaves a critical gap in systematically assessing how MLLMs withstand real-world manipulations.

To bridge this gap, we introduce \textbf{MVTamperBench}, a benchmark specifically designed to evaluate MLLM robustness against five prevalent tampering techniques. Built from 3.4K original videos—expanded into over 17K tampered clips spanning 19 video tasks—MVTamperBench challenges models to detect manipulations by disrupting temporal and spatial coherence in a video. By focusing on tampering resilience, we believe that {MVTamperBench} provides a critical tool for improving MLLM robustness, and eventually, it enables the development of tamper-resilient models applicable to real-world challenges such as clickbait detection, content moderation, and policy enforcement, advancing adversarial robustness in video understanding for high-stakes domains.


Our main contributions are as follows: 
\begin{itemize}
    \item We introduce \textbf{MVTamperBench}, a benchmark that systematically evaluates MLLMs on five major video manipulations, focusing on spatial and temporal coherence to stimulate real-world scenarios.
    \item We propose a unified evaluation methodology that frames tampering detection as a multiple-choice task, enabling straightforward, interpretable and consistent performance comparisons.
    \item Through experiments on 45 MLLMs across 15+ families, we identify critical vulnerabilities across MLLM families.
    across MLLM families, without any correlation between model size and performance. 
    \item Our released code enables researchers to integrate additional datasets and adapt or add new tampering, facilitating domain-specific extensions, and supports reproducibility.

\end{itemize}

\section{Related Work}

The development of benchmarks for MLLMs has significantly advanced the evaluation of image and video understanding tasks. They have covered spatial reasoning, temporal comprehension, object detection, common sense inference, and so on. However, the robustness of MLLMs to adversarial manipulations, particularly video tampering, remains underexplored. This section reviews existing benchmarks, which can be categorized into image-based and video-based evaluations, and analyzes their contributions and shared limitations.

\subsection{Image-based Understanding}

Benchmarks like BLINK \cite{Fu:25} and MuirBench \cite{Wang:24} focus on evaluating static visual reasoning. BLINK tests foundational tasks such as depth estimation, forensic detection, and visual correspondence, which cover a diverse set of challenges for spatial reasoning. Similarly, MuirBench extends this evaluation to multi-image tasks, including action recognition and geographic reasoning, by synthesizing information from diverse sources. While these benchmarks have advanced static image understanding, their reliance on single or multiple still images excludes temporal dynamics and limits their applicability to scenarios involving sequential manipulations, such as those found in video content.

\subsection{Video-based Understanding}

Video-based benchmarks have expanded the scope of evaluation by incorporating temporal and multimodal reasoning. MVBench \cite{Li:24}, MMBench-Video \cite{Fang:24}, and Video-MME \cite{Fu:24-VideoMME} focus on tasks such as event detection, episodic reasoning, and contextual understanding. These benchmarks challenge models with diverse tasks spanning object interactions, long-duration video analysis, and domain-specific reasoning. Similarly, LongVU \cite{Shen:24-LongVU} introduces spatio-temporal compression techniques to enhance efficiency, while MotionEpic \cite{Fei:24} integrates Spatial-Temporal Scene Graphs (STSG) for fine-grained cognitive tasks. However, while these benchmarks assess temporal coherence and reasoning, they do not systematically address adversarial robustness, particularly in tampering scenarios.

\begin{figure*}[th!]
    \centering
    \includegraphics[width=\textwidth]{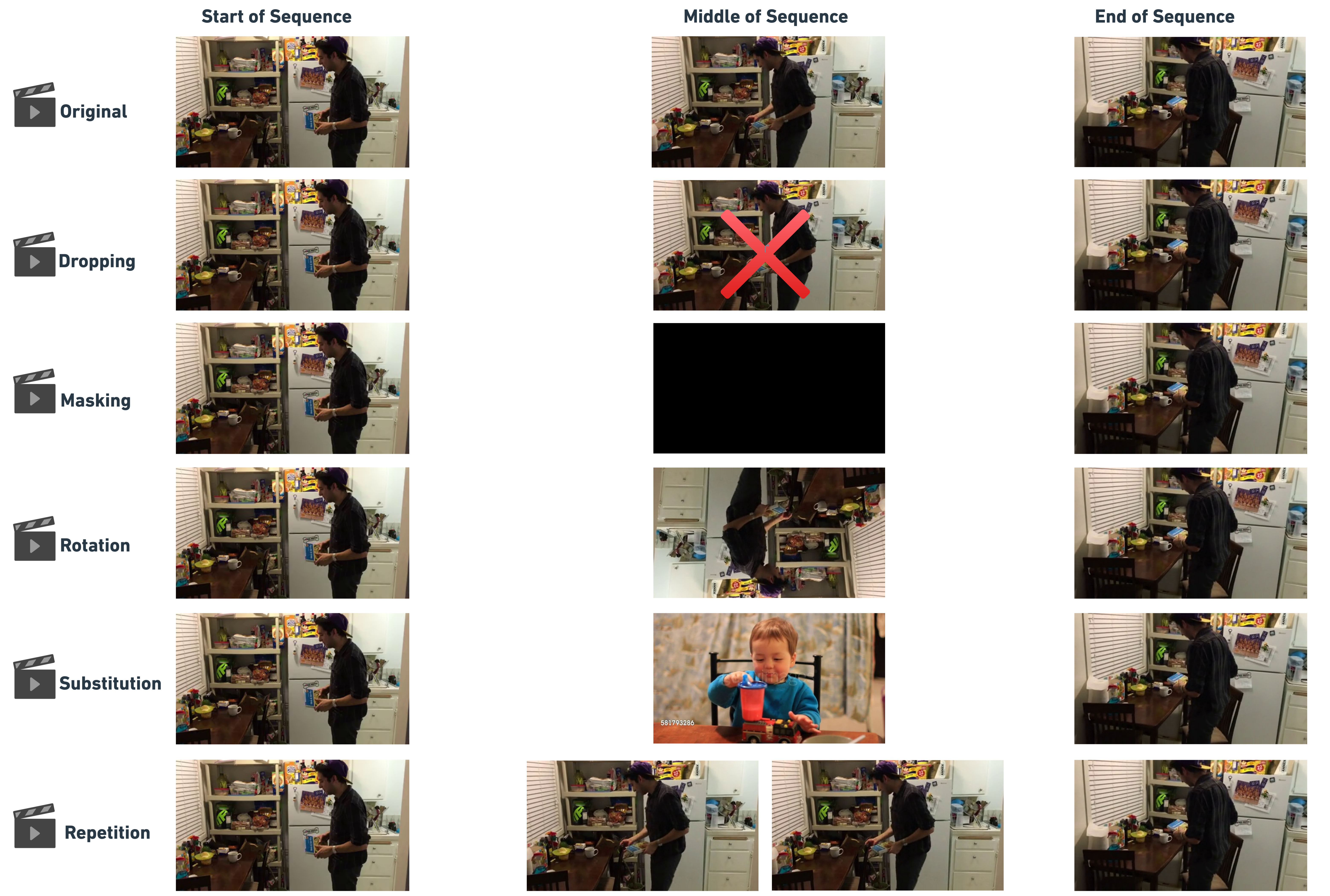}
    \caption{{Illustration of the five video frame tampering techniques.} Each row shows a specific tamper type applied to a video snippet, describing the respective impact on temporal/spatial coherence.}
    \label{fig:tampered_frame_image}
    \vspace{-1.0em}
\end{figure*}

Other benchmarks, such as Wolf \cite{Li:24-wolf} and Sharingan \cite{Chen:24}, target specialized video understanding tasks. Wolf focuses on captioning using a mixture-of-experts strategy, while Sharingan extracts action sequences from desktop recordings using frame-differential approaches. Although these benchmarks achieve high accuracy in their domains, they are limited to specific tasks and lack general mechanisms to evaluate the resilience to tampering.

In summary, despite their contributions to advancing MLLM evaluation, existing benchmarks largely focus on performance under ideal conditions and neglect robustness to tampering or adversarial effects. Techniques such as frame substitution, masking, repetition, dropping, and rotation disrupt temporal coherence and pose unique challenges for multimodal models. The absence of systematic evaluations of these tampering techniques highlights a critical gap in ensuring model reliability for real-world applications like forensic analysis, media verification, and misinformation detection.

\section{MVTamperBench}
\label{sec:mvtamperbench}
We introduce \textbf{MVTamperBench}, a comprehensive benchmark designed to systematically evaluate the robustness of MLLMs against video tampering techniques. Through our MVTamperBench which introduces diverse manipulations, we aim to broaden the evaluation landscape, enabling a deeper understanding of model strengths and vulnerabilities under adversarial scenarios. Table \ref{tab:benchmark_comparison} 
compares MVTamperBench with existing benchmarks, highlighting its focus areas, strengths, and unique contributions. 

In the following subsections, we detail its construction, design choices, and key features. More details can be found in our code\footnote{https://amitbcp.github.io/MVTamperBench/}, data\footnote{https://hf.co/datasets/Srikant86/MVTamperBench} and benchmark \footnote{https://github.com/open-compass/VLMEvalKit} repositories.



\subsection{Benchmark Construction}

To evaluate MLLM robustness under adversarial conditions, we apply distinct tampering methods as shown in Figure \ref{fig:tampered_frame_image} —\emph{Dropping}, \emph{Masking}, \emph{Substitution}, \emph{Repetition}, and \emph{Rotation}—to the 3,487 original MVBench \cite{Li:24} videos (excluding NTU dataset due to licensing), resulting in a total of 17,435 tampered clips. These manipulations target both spatial and temporal coherence, thereby simulating common real-world tampering scenarios such as deliberate frame editing or slicing from unrelated content.

\subsection{Tampering Techniques}
\noindent \textbf{Dropping:} Removes a 1-second segment for creating temporal discontinuity.\\
\noindent \textbf{Masking:} Overlays a black rectangle on a 1-second segment. It aims to simulate visual data loss.\\
\noindent \textbf{Rotation:} Rotates a 1-second segment by 180 degrees for introducing spatial distortion.\\
\noindent \textbf{Substitution:} Replaces a 1-second segment with a pre-selected clip from another video, in order to disrupt temporal and contextual flow.\\
\noindent \textbf{Repetition:} Repeats a 1-second segment, introducing temporal redundancy. 

\noindent The aforementioned effects are applied uniformly across all videos to ensure consistent and comparable evaluation.

\subsection{Design \& Implementation }
\paragraph{Tampering Duration (1 Second).} We fix tampering to a 1-second segment to align with reports of minimal but impactful real-world tampering, e.g., short edits on social networks or subtle modifications on surveillance feeds \cite{reolink2025tampering}. Our preliminary experiments revealed that using less than 1 second could be overlooked by certain model sampling mechanisms, whereas longer tampering (>3s) often resembled normal scene transitions, reducing adversarial impact.

\paragraph{Tampering Location (Middle).} All manipulations occur at the video’s midpoint to disrupt central content. Our pilot tests showed that tampering near the start or end risked mimicking scene cuts or information loss, which makes detection less indicative of genuine adversarial robustness.

\paragraph{Substitution Source.} For Substitution, the 1-second clip is randomly chosen from a consistent pool of different videos within MVBench. This ensures uniform difficulty across all samples, thus preventing confounds from domain shifts or overly simplistic substitutes.

\paragraph{Modular \& Scalable Framework.} Each technique is encapsulated in a reusable class for facilitating custom parameterization (e.g., tampering duration, location, intensity). Our open-source code will integrate seamlessly with \textbf{VLMEvalKit}~\cite{duan2024vlmevalkit}, thereby promoting reproducible experiments and easy extension to other datasets or tasks.

To inform our final design decisions, we conducted a series of exploratory experiments examining the effects of tampering position and duration. Results from these alternative design configurations are detailed in Appendix \ref{sec:appendix-design}.

\subsection{Dataset Scope and Statistics}
\begin{figure}[h!]
\centering
\includegraphics[width=0.48\textwidth]{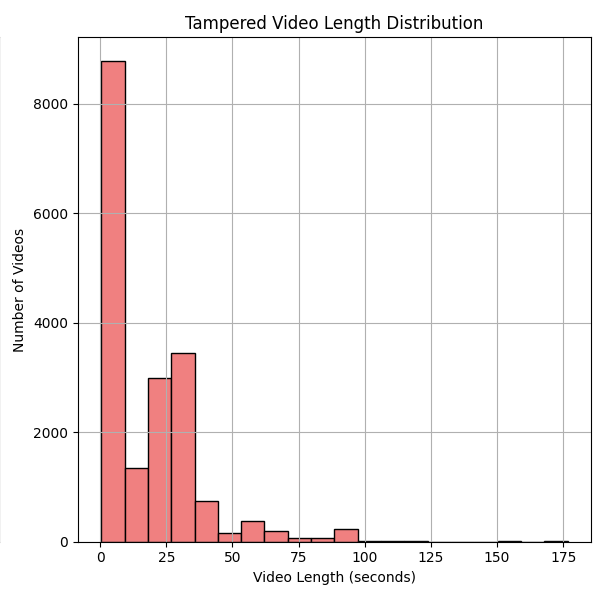}
\caption{Distribution of video durations. Our dataset spans a broad range of durations, which can reflect varied real-world conditions.}
\label{fig:Video Length Distribution}
\vspace{-1.5em}
\end{figure}

All 3,487 MVBench videos undergo each of the five tampering types, producing 17,435 tampered clips. The five manipulations are uniformly applied to ensure comparability across different MLLM architectures and training regimes. Figure~\ref{fig:Video Length Distribution} illustrates the distribution of their durations. We can observed that diverse scenarios from short (3-5s) to extended (>20s) sequences are included. 

\subsection{Summary}
By enforcing consistent parameters (1s duration, midpoint placement) and systematically applying five tampering methods, \textbf{MVTamperBench} offers a controlled yet flexible platform for evaluating tampering resilience. Our design can be easily extended with additional manipulations (e.g., noise injection, partial masking, positions), deepfakes or integrated into domain-specific contexts like surveillance feeds analysis or clickbait detection. We provide additional details on MVTamperBench, including video sources and associated tasks, in Appendix \ref{sec:appendix_data}.


\begin{figure*}[th!]
    \centering
    \includegraphics[width=1\textwidth]{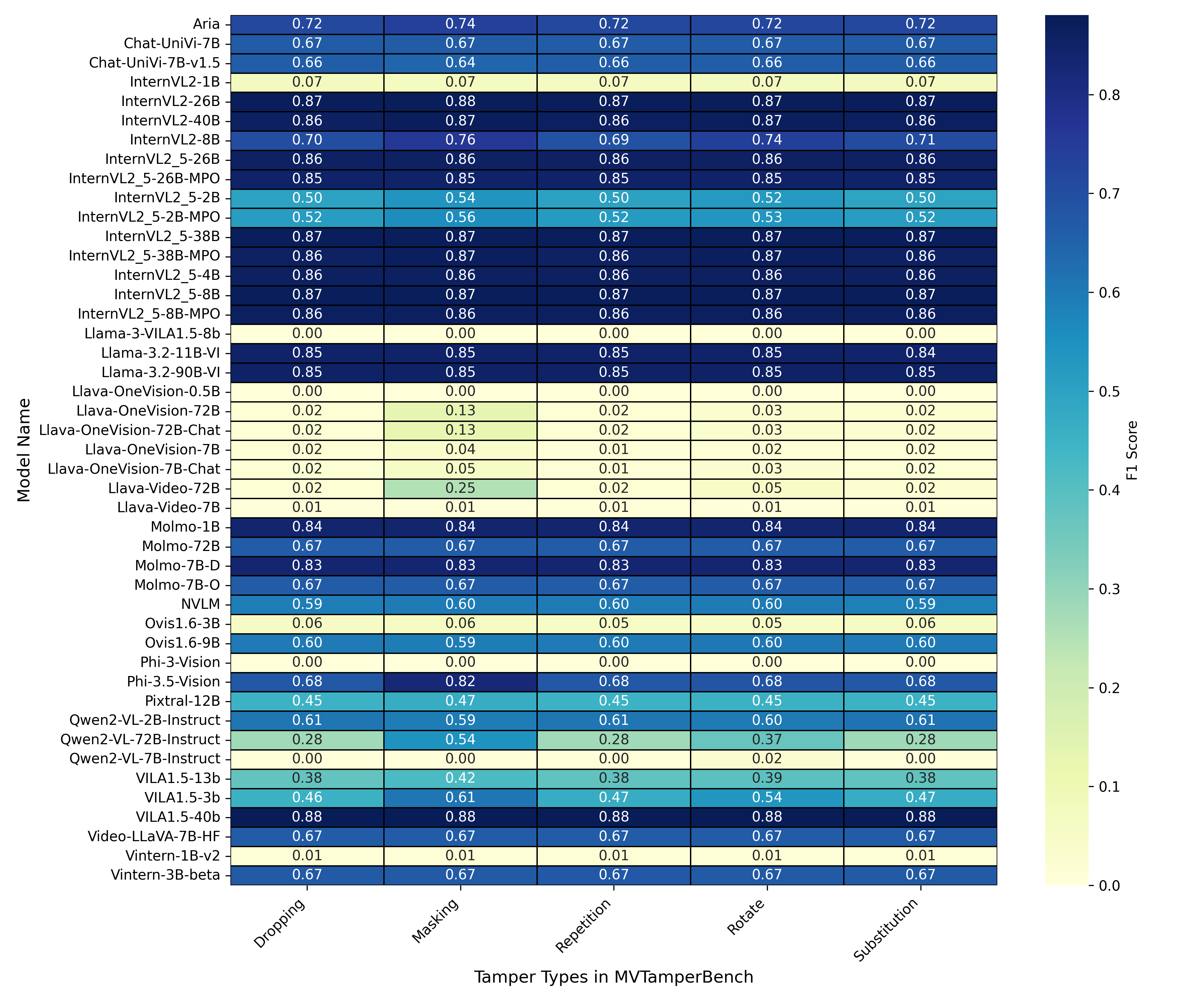}
    \caption{F1 scores across models and tampering types. High-performing models are robust across all types.}
    \vspace{-1.1em}
    \label{fig:heatmap}
\end{figure*}


\section{Experiments}

This section outlines the experimental setup, results, and analysis of 45 models evaluated on our proposed benchmark. Our results highlight their strengths, weaknesses, and actionable insights. We provide further overview of the evaluated MLLMs in Appendix \ref{sec:appendix_overview_lmm}.

\subsection{Experimental Setups}


\paragraph{Evaluation Protocol} Each model is tasked with identifying whether a video has been tampered with or not. For every video, the model is presented with the following structured prompt: 
\vspace{-0.3em}
\begin{quote}
    \emph{Does this video exhibit any signs of tampering, such as corruption, blackouts, rotated frames, repeated frames, or swapped frames?}\\
    \textbf{Options:} A. Yes \hspace{1em} B. No
\end{quote}
\vspace{-0.3em}


The dataset comprises both tampered and non-tampered videos. For each tampered video, the corresponding non-tampered video is included to ensure a balanced distribution. Models must select one of the two options (\textbf{Yes} or \textbf{No}), and their predictions are compared against ground truth labels to determine correctness. This task is repeated for all tampering types.

We also compared structured, general, and chain-of-thought prompts (Appendix \ref{sec:appendix-prompt}), finding that general and CoT variants led to higher false positives due to limited temporal reasoning in current MLLMs. 

\paragraph{Metrics} The primary evaluation metric is the \textbf{F1 Score}, chosen for its ability to balance precision and recall, particularly in scenarios where misclassifications (false positives and false negatives) can significantly impact robustness evaluation. We believe that this metric is the most suitable for our setup, where models must not only detect manipulations but also avoid false positives on non-tampered videos. To capture performance across all tampering types, we compute individual F1 scores for each tampering type. In addition, we also measure \textbf{F1 (overall) score} as the macro average of these individual F1 scores. We thus highlight per-tamper-type strengths and weaknesses as well as overall model performance.


\begin{figure}[th!]
    \centering
    \includegraphics[width=0.5\textwidth]{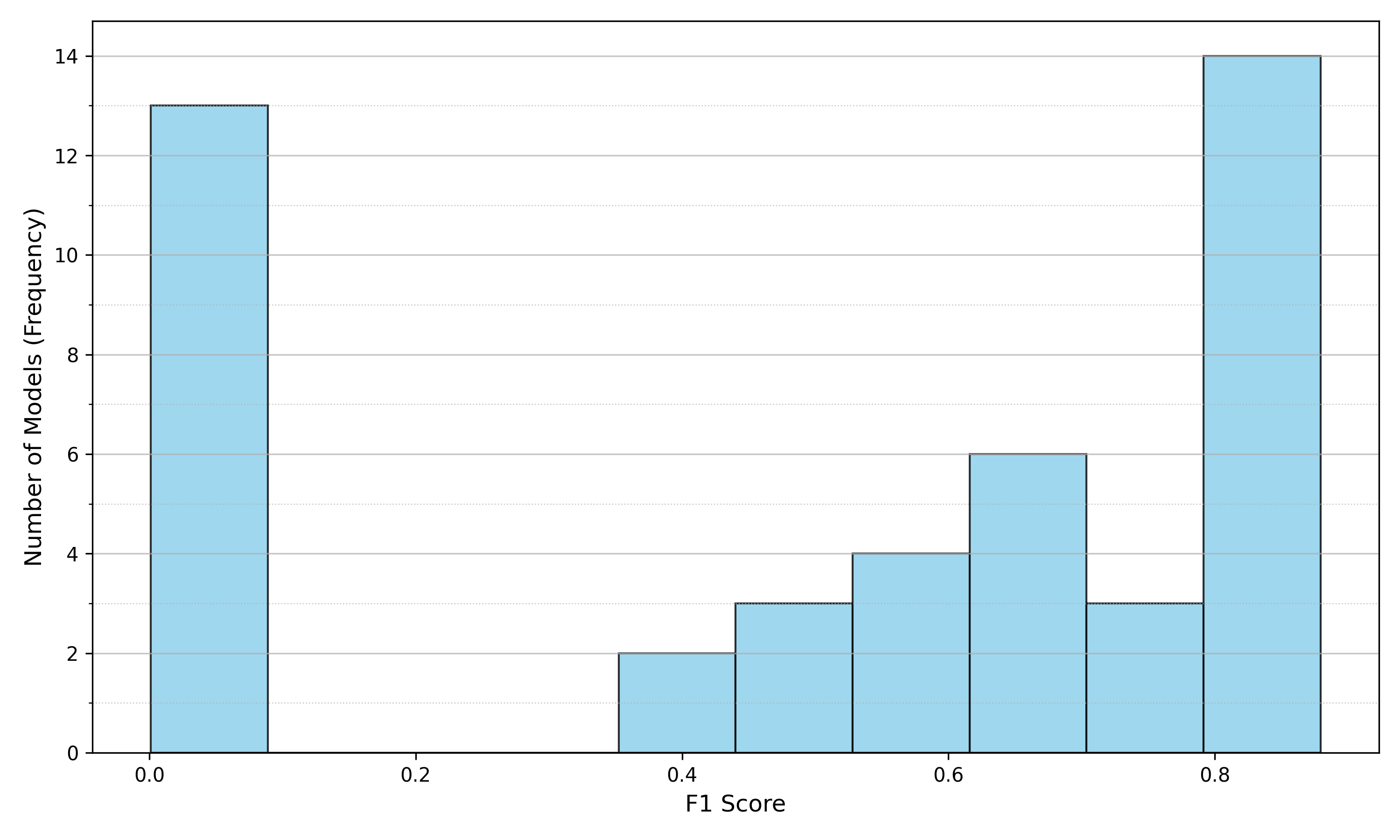}
    \caption{Distribution of F1 (overall) scores across models. 
    }
    \vspace{-1.5em}
    \label{fig:f1_dist_across_model}
\end{figure}

\begin{figure}[th!]
    \centering
    \includegraphics[width=0.5\textwidth]{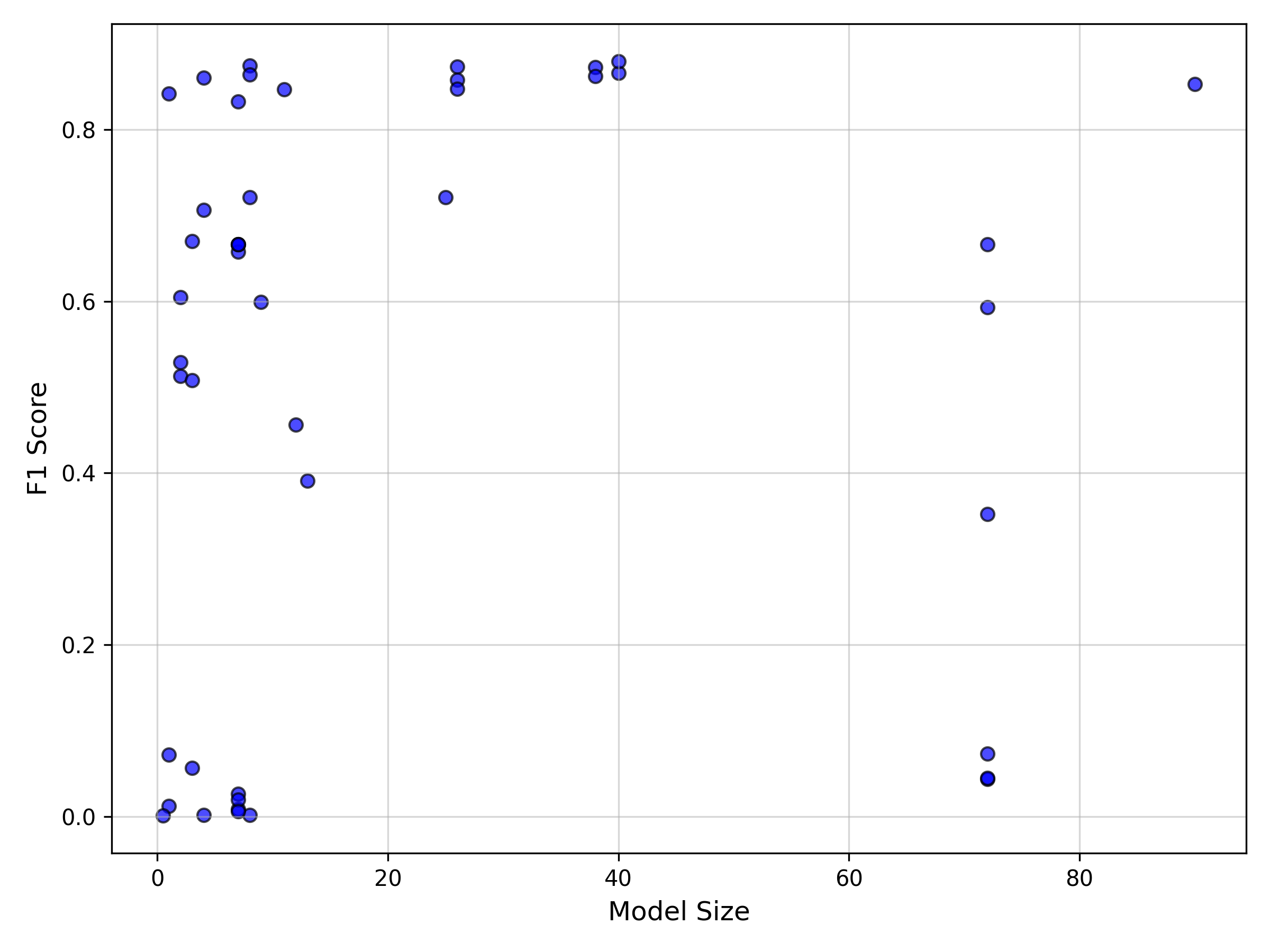}
    \caption{Scatter plot showing no correlation between model size and overall F1 (Pearson r=0.05).
    }
    \vspace{-1.5em}
    \label{fig:size_corr}
\end{figure}

\subsection{Results and Analysis}

We evaluate the performance of 45 MLLMs using F1 (overall) scores and individual F1 scores across five tampering types. The distribution of F1 (overall) scores (Figure \ref{fig:f1_dist_across_model}) reveals significant variability in the robustness of the model, with several models struggling to detect tampering (F1 < 0.2), while a few high-performing models achieve F1 > 0.8. 

We did not observe a correlation between model size and tampering detection performance (Pearson correlation = 0.05, Figure \ref{fig:size_corr}). This highlights that architectural differences and training techniques, rather than parameter count, contribute more significantly to robustness against tampering.

In addition, Figure \ref{fig:heatmap} showcases model-specific adaptability across different tampering types. Models with F1 (overall) > 0.8 are consistently robust across all tampering types, while those with F1 < 0.2 perform slightly better on \textit{Masking}, which relies more on spatial reasoning. Models in between generally struggle with temporal disruptions like \textit{Dropping} and \textit{Substitution}, reflecting challenges in temporal coherence.



\subsubsection{Analysis based on Performance}

\begin{figure}[th!]
    \centering
    \includegraphics[width=0.5\textwidth]{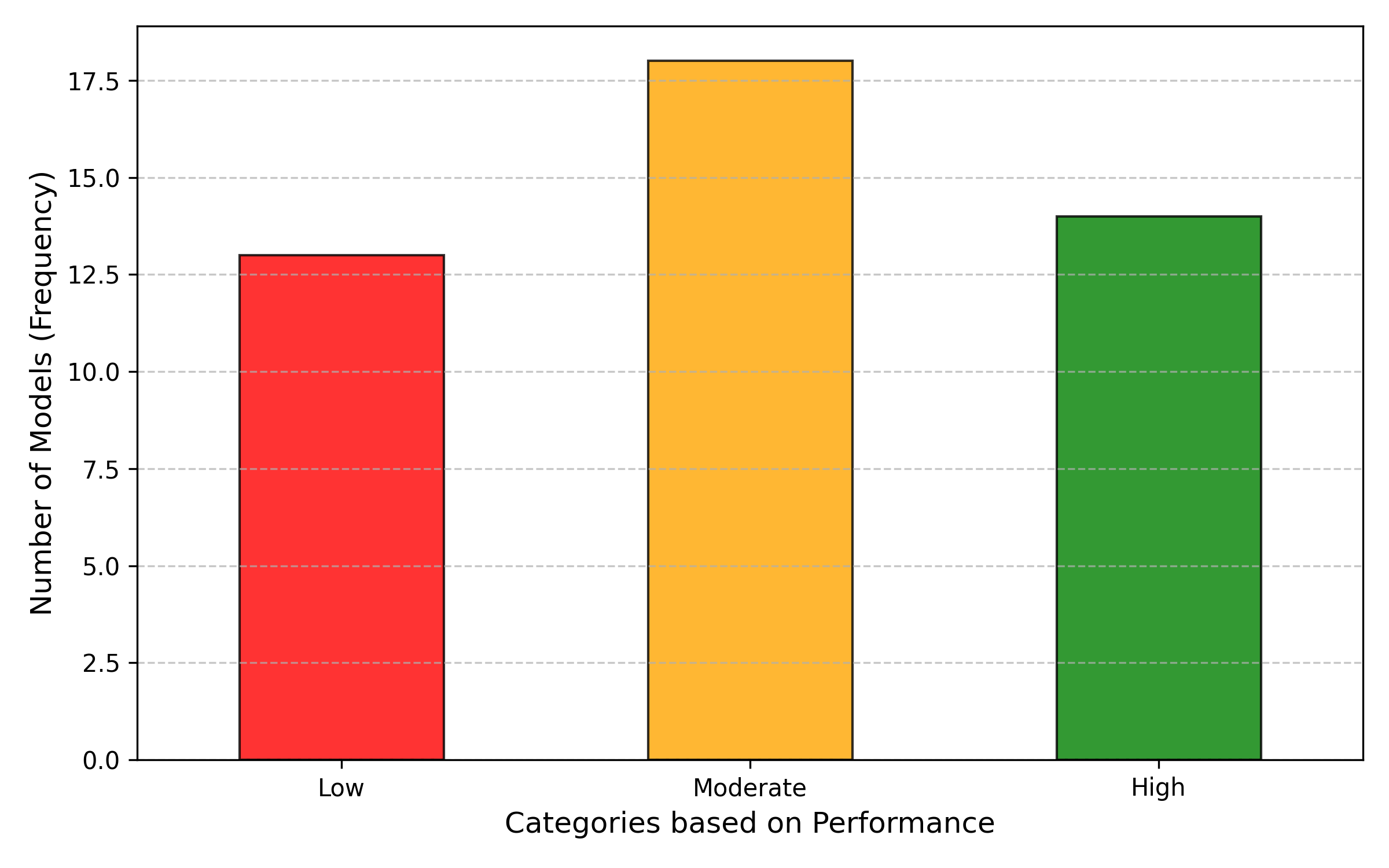}
    \caption{Distribution of Number of Models in low, moderate, and high-performing categories based on F1 (overall).
    }
    \label{fig:model_freq_perf_cat}
\end{figure}

Figure \ref{fig:model_freq_perf_cat} categorizes MLLMs into \textbf{low-}, \textbf{moderate-}, and \textbf{high-} performing groups based on their F1-score distribution. We define the boundaries using the 0.25 quantile (F1 = 0.071) and 0.75 quantile (F1 = 0.846). We round-off the quantile thresholds to the nearest integers to identify low-performing models (F1 (overall) < 0.01) and high-performing models (F1 (overall) > 0.8).

\begin{figure}[th!]
    \centering
    \includegraphics[width=0.5\textwidth]{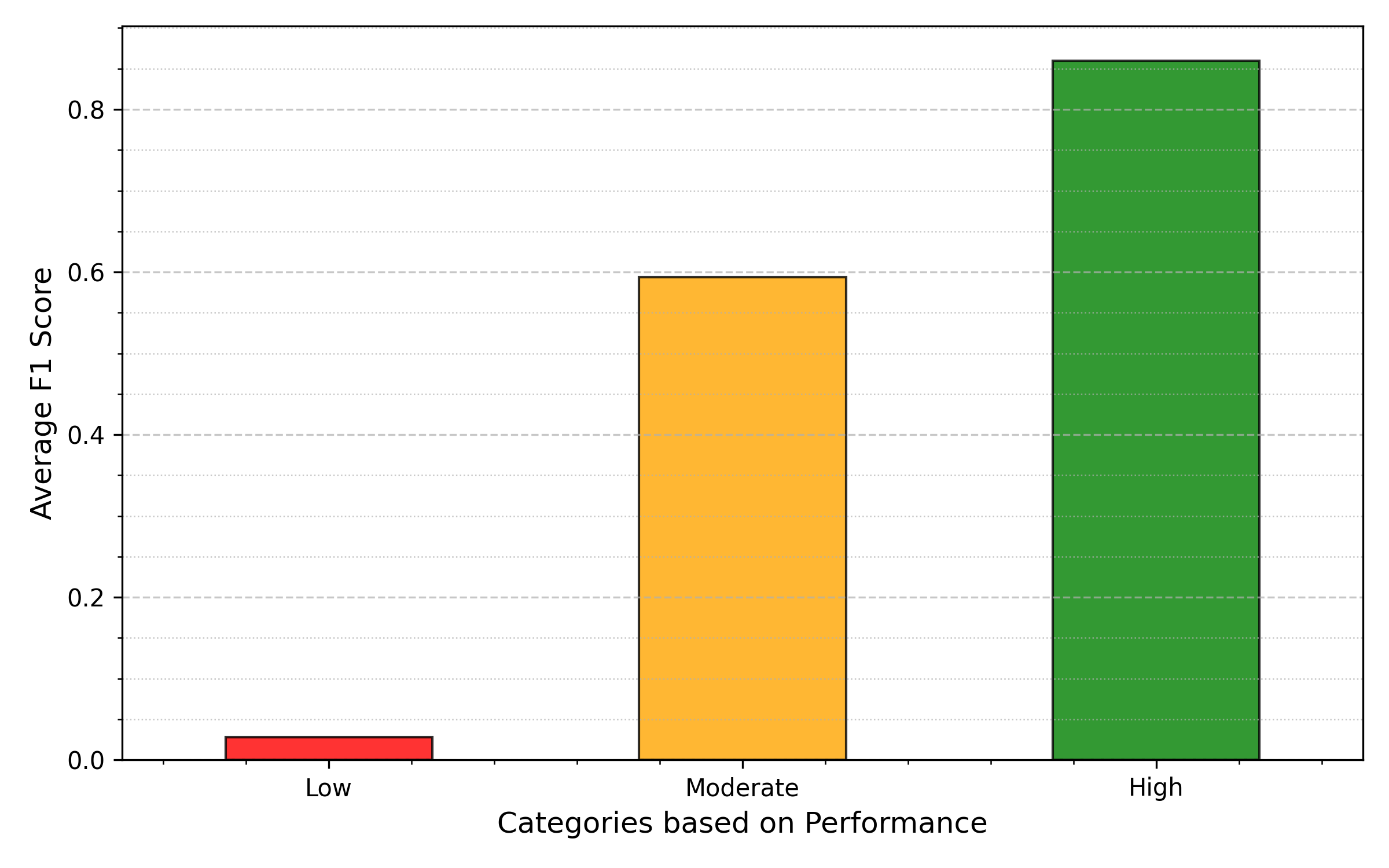}
    \vspace{-1.5em}
    \caption{Average F1 (overall) scores across low, moderate, and high-performing model categories.}
    
    \label{fig:f1_perf_cat}
\end{figure}

\begin{figure}[th!]
    \centering
    \includegraphics[width=0.5\textwidth]{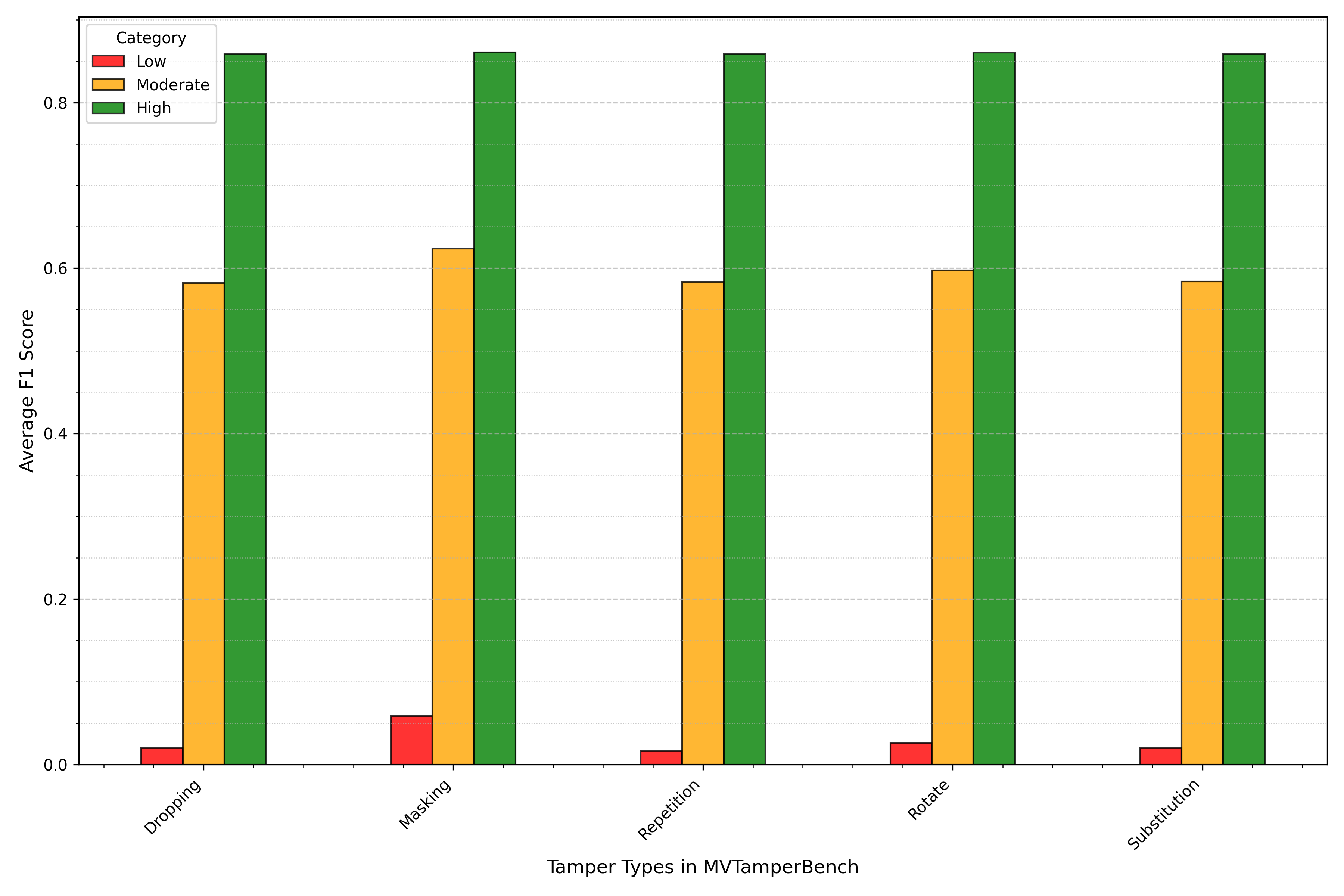}
    \caption{F1 scores for tampering types across model categories. Low \& Moderate-performing models perform slightly better on \textit{Masking}, while high-performing models show consistent robustness.}
    \label{fig:task_perf_cat}
\end{figure}

Figure \ref{fig:f1_perf_cat} illustrates the average F1 scores for each category. We observe stark differences between groups: High-performing models maintain consistent performance across all tampering types, whereas low-performing models show significant weaknesses, particularly for tampering effects that disrupt temporal coherence (\textit{Dropping}, \textit{Substitution}). Moderate-performing models excel slightly in \textit{Masking} (Figure \ref{fig:task_perf_cat}).

\begin{figure}[th!]
    \centering
    \includegraphics[width=0.5\textwidth]{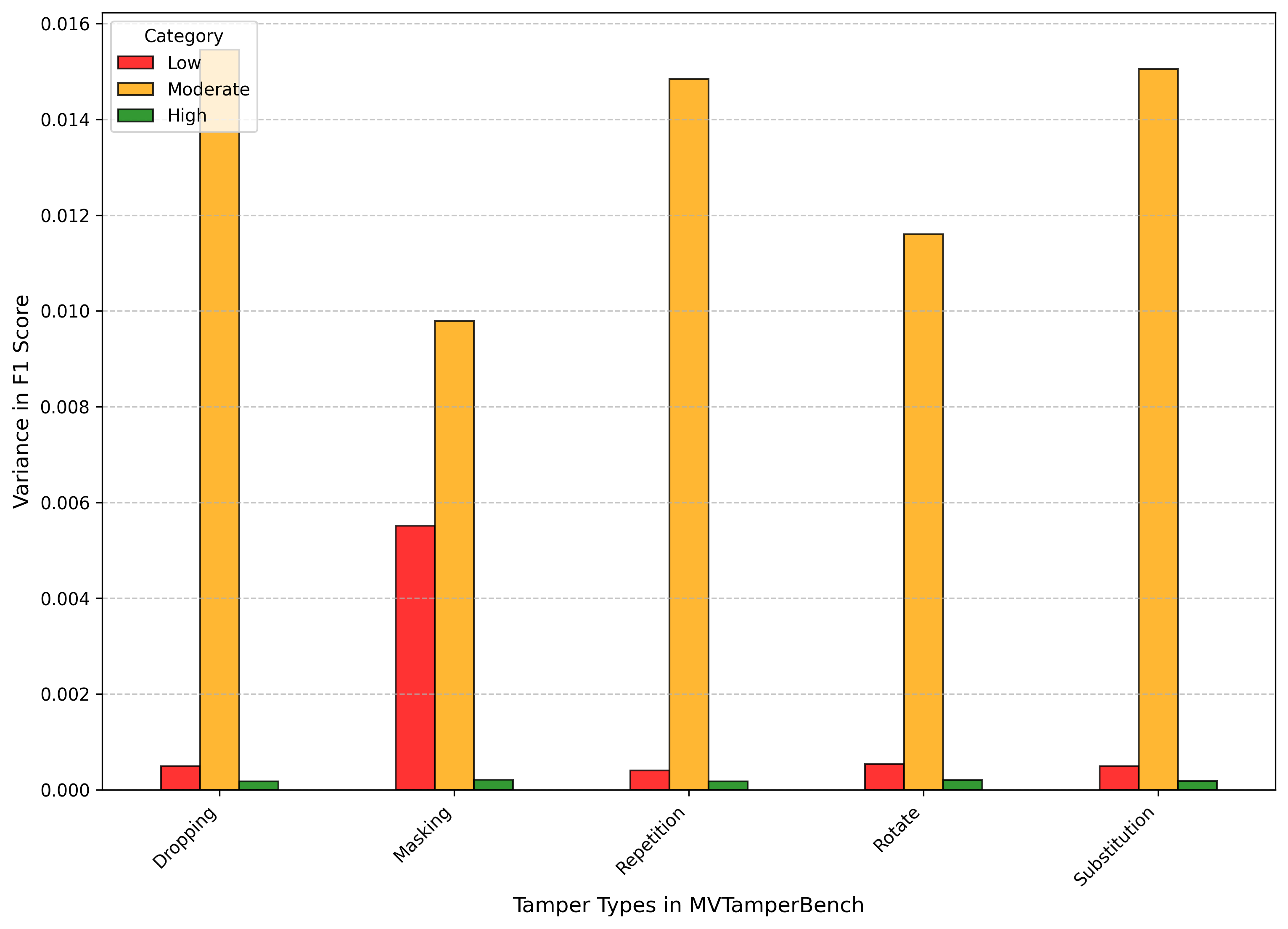}
    \caption{Variance in F1 scores across tampering types for each model category. 
    }
    \vspace{-1.5em}
    \label{fig:task_perf_cat_variance}
\end{figure}




We analyze the variance in model performance between tampering types in Figure \ref{fig:task_perf_cat_variance}. High-performing models exhibit negligible variance, indicating consistent robustness across all tampering effects. This stability suggests that these models are well-equipped to handle both spatial \& temporal disruptions introduced by tampering.

Low-performing models display minimal variance across most tampering types, with the exception of \textit{Masking}, which shows a slightly higher variance. This highlights a specific weakness in processing visual obfuscations, likely due to their reliance on static features rather than robust temporal reasoning. The minimal variance across other tampering types suggests that these models fail uniformly, regardless of the type of manipulation.

Moderate-performing models demonstrate the \textbf{highest variance overall}, particularly for tampering types such as \textit{Dropping}, \textit{Repetition}, and \textit{Substitution}. This behavior indicates an inconsistency in their ability to adapt to tampering effects. While these models achieve a balanced performance across easier tampering types, temporal disruptions like \textit{Dropping}, \textit{Repetition}, and \textit{Substitution} pose significant challenges, leading to occasional spikes in variance. This suggests that moderate-performing models, while more capable than low-performing counterparts, still struggle to maintain robustness across a diverse set of tampering scenarios. We provide a detailed analysis of individual model performance in Appendix~\ref{sec:appendix_model_perf}, and present qualitative trends based on model architecture and training paradigms across all models studied in Appendix~\ref{sec:appendix-architecture-trends}, offering insights into the factors influencing performance.

\subsubsection{Analysis based on Model Size}

\begin{figure}[H]
    \centering
    \includegraphics[width=0.5\textwidth]{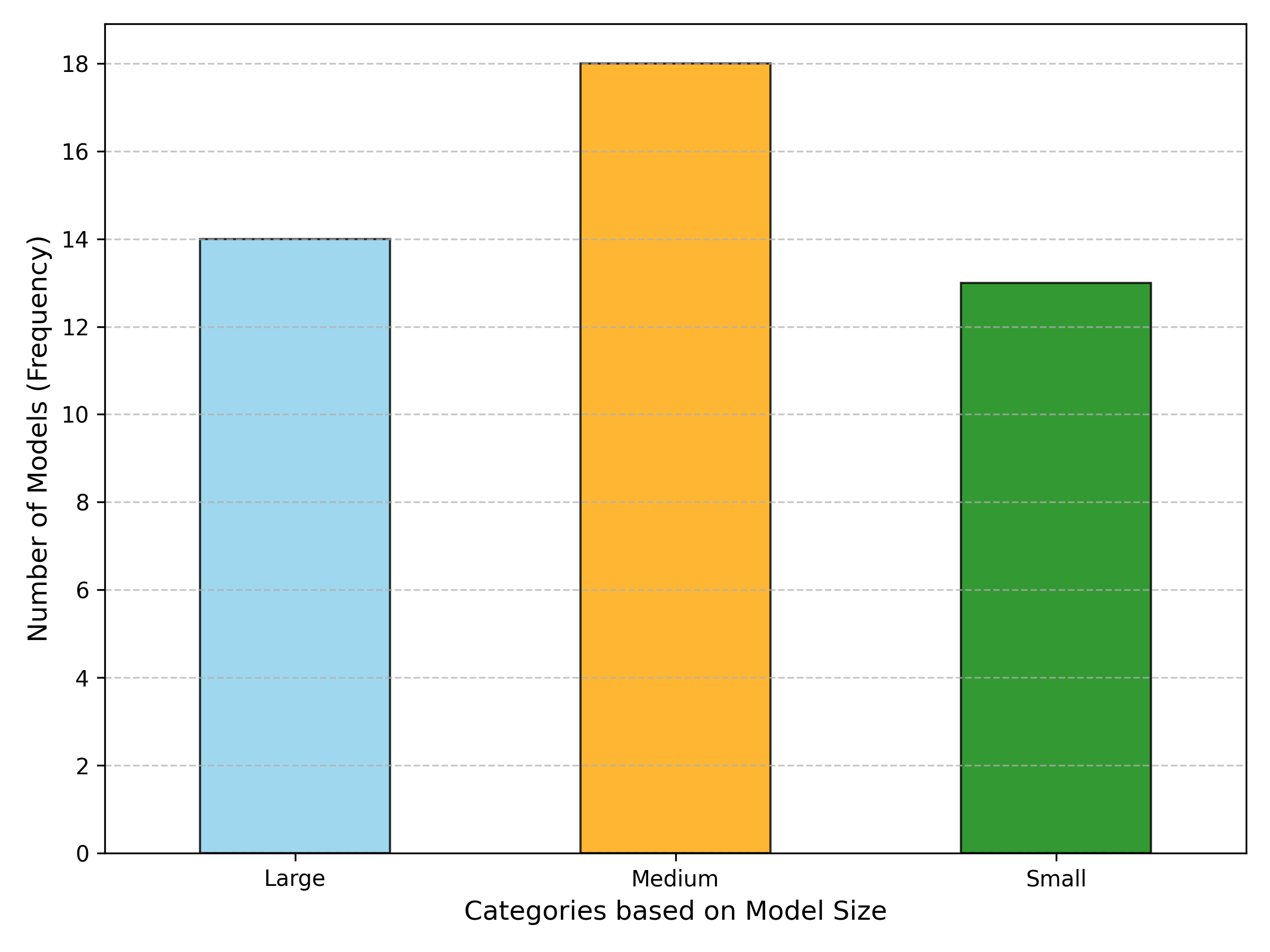}
    \caption{Distribution of models grouped by size; Categories: Small (<7B), Medium (7B–26B), and Large (>26B).}
    \vspace{-2em}
    \label{fig:model_freq_model_size}
\end{figure}

\begin{figure}[H]
    \centering
    \includegraphics[width=0.5\textwidth]{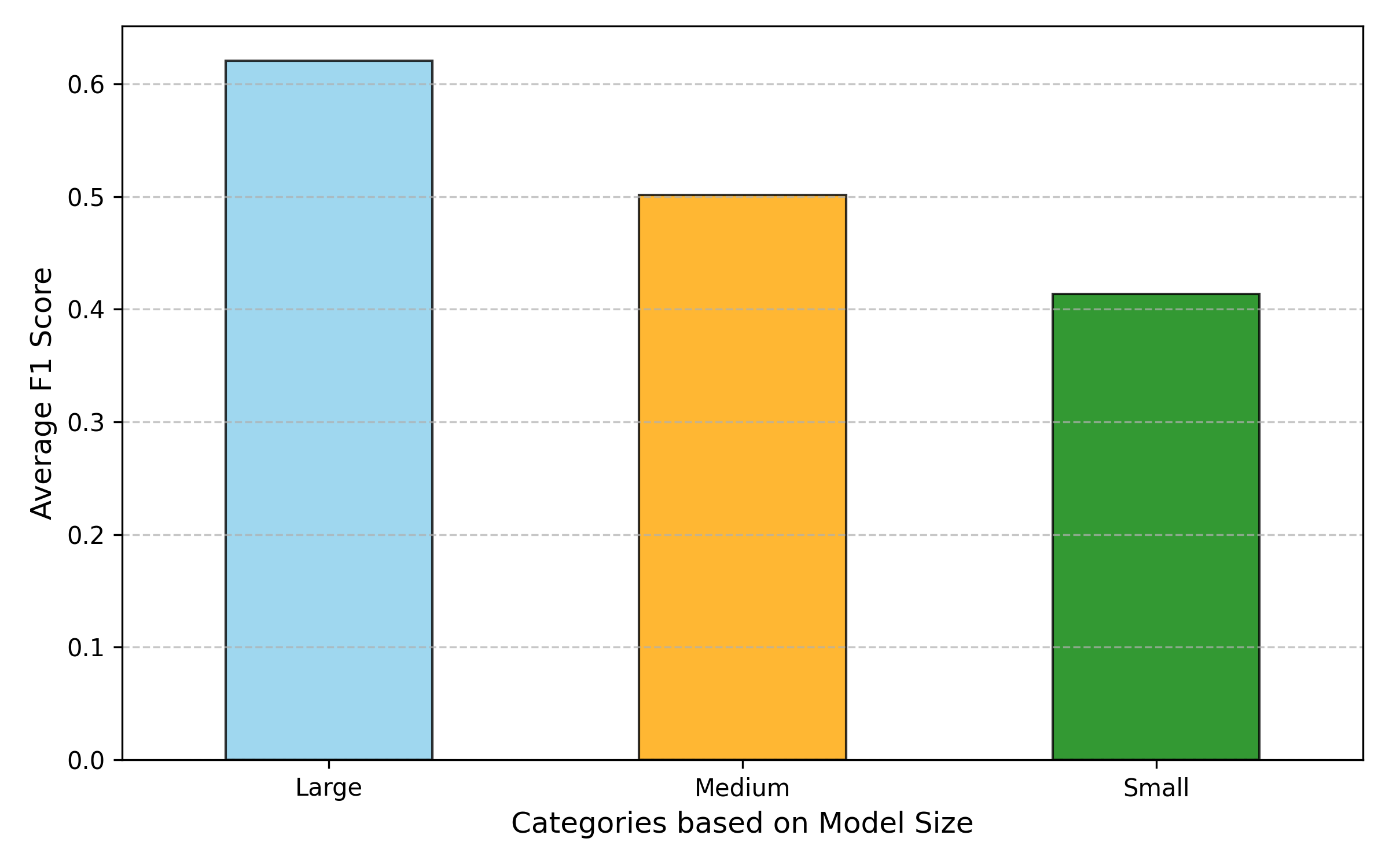}
    \caption{F1 (overall) scores for models by size category. Larger models generally tend to perform better.}
    \label{fig:f1_by_model_size}
\end{figure}

We categorize models based on their parameter sizes into \textbf{small} (<7B), \textbf{medium} (7B–26B), and \textbf{large} (>26B) groups (Figure \ref{fig:model_freq_model_size}). While Figure \ref{fig:f1_by_model_size} shows that larger models generally achieve higher F1 (overall) scores, Figure \ref{fig:size_corr} confirms no significant correlation between model size and tampering detection performance (Pearson correlation = 0.05). A closer examination of individual model trends across size categories reveals several noteworthy patterns discussed in Appendix \ref{sec:appendix_model_size}.

\begin{figure*}[th!]
    \centering
    \includegraphics[width=0.95\textwidth]{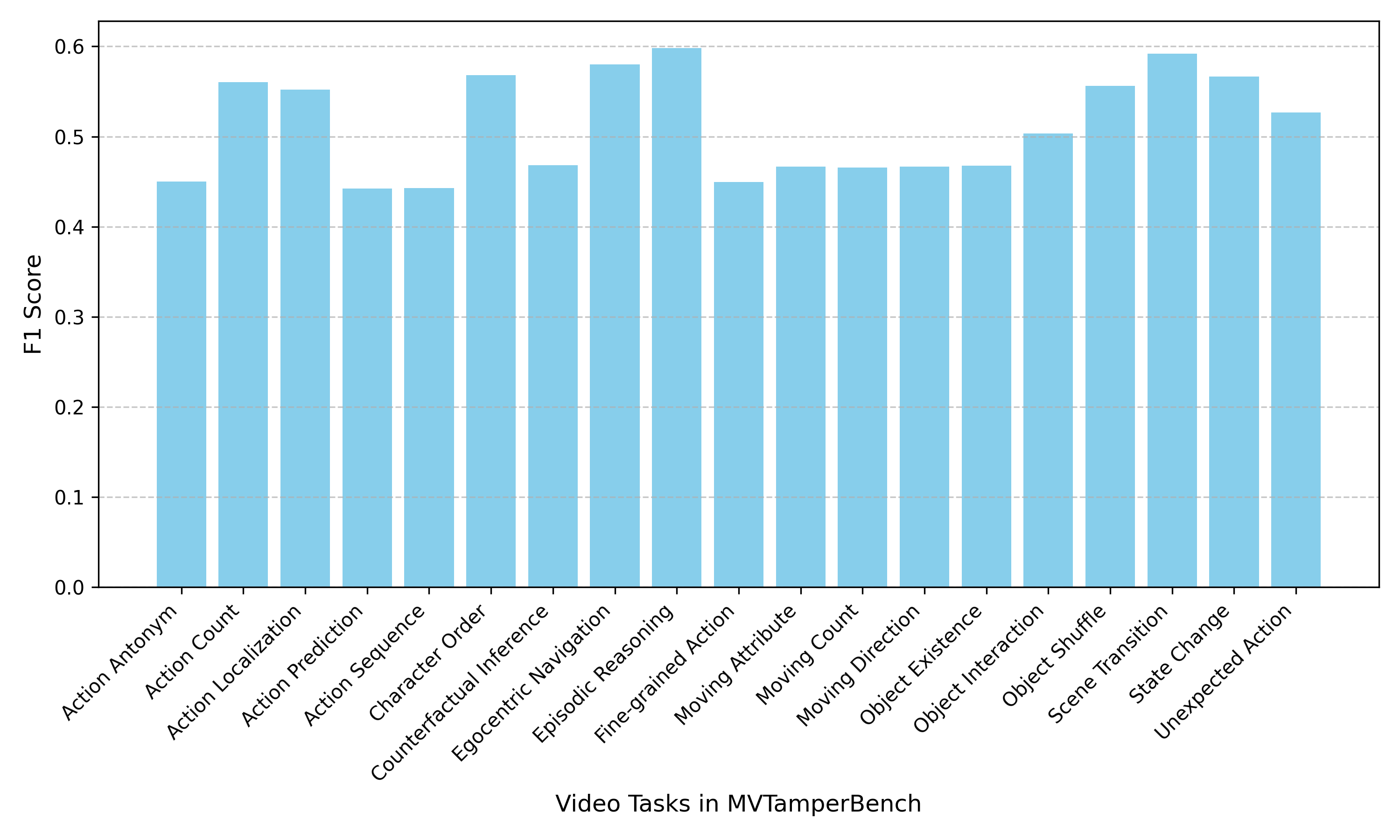}
    \vspace{-3mm}
    \caption{F1 (overall) scores for task categories in MVTamperBench. Tasks like \textit{Episodic Reasoning} achieve higher scores, while \textit{Counterfactual Inference} is more challenging.}
    \vspace{-1.5em}
    \label{fig:f1_by_task_type}
\end{figure*}


 

\paragraph{Trends Across Families.} Across size categories, we observe distinct trends in model performance. The \textbf{VILA model family} consistently improves with size, with the exception of \textbf{VILA1.5-8B}, highlighting the scalability of its architecture for tampering detection. Similarly, \textbf{Qwen2-VL} demonstrates significant gains with increased parameters, though it trails behind other families in absolute performance.

The \textbf{Llama3.2-Vision} family, despite its scaling efforts, reveals the diminishing returns of increasing model size without architectural or training advancements. Meanwhile, \textbf{Molmo} models illustrate the importance of efficient design, as \textbf{Molmo-1B} outperforms its larger variants like \textbf{Molmo-72B}. Finally, the dominance of \textbf{InternVL-2.5} across all sizes highlights the benefits of balanced architecture \& task-specific training strategies.

\vspace{-0.2em}
\subsubsection{Analysis across Video Task Types}

MVTamperBench comprises 19 video task categories, each evaluated across five tampering techniques. Figure \ref{fig:f1_by_task_type} highlights the F1 (overall) scores for each task, averaged across all tampering types and models. While certain tasks exhibit high detection F1 score, others remain significantly more challenging. Appendix \ref{sec:appendix_video_task} provides additional insights into the performance trends of each tampering type across different tasks.


\paragraph{Easier Tasks.}
Tasks such as \textbf{Episodic Reasoning, Scene Transition, Ego-centric Navigation,} and \textbf{State Change} consistently achieve higher F1 scores, as shown in Figure \ref{fig:f1_by_task_type}. These tasks often involve shorter temporal dependencies and simpler spatial reasoning, allowing models to rely on pre-trained vision-language features rather than advanced temporal reasoning.


For example, in \textbf{Episodic Reasoning}, tampering effects like \textit{Dropping} and \textit{Repetition} minimally affect task performance, as models primarily focus on memorizing what it has seen. Similarly, \textbf{Scene Transitions} tasks depend more on detecting static or localized changes, making them less susceptible to temporal disruptions such as \textit{Dropping} or \textit{Repetition}. The results indicate that models trained on extensive image-based datasets excel in tasks with lower temporal complexity.


\paragraph{Challenging Tasks.} Tasks such as \textbf{Action Prediction, Counterfactual Inference,} and \textbf{Fine-Grained Action} pose significant challenges for tampering detection. These tasks inherently require complex temporal reasoning and context preservation, making them highly sensitive to disruptions introduced by tampering techniques.

In \textbf{Counterfactual Inference}, tampering effects like \textit{Substitution} \& \textit{Dropping} cause significant confusion, as they disrupt the continuity of the video narrative required for hypothetical reasoning. Similarly, \textbf{Fine-Grained Action} detection is heavily impacted by \textit{Rotation}, distorting spatial relationships crucial for identifying subtle movements.


Interestingly, \textbf{Action Prediction} tasks also highlight the limitations of current MLLMs in understanding and reasoning about temporal progression. Models often fail to account for disruptions in video sequences, leading to degraded performance across tampering types.

We further discuss our Key Findings \& Future Directions in Appendix \ref{sec:appendix_key} \& \ref{sec:appendix_discussion} respectively.

\section{Concluding Remarks}

\paragraph{Novel Benchmark.} We introduced {MVTamperBench}, a comprehensive benchmark for evaluating the robustness of Multimodal Large Language Models (MLLMs) against five key video tampering techniques—\emph{Dropping}, \emph{Masking}, \emph{Repetition}, \emph{Rotation}, and \emph{Substitution}. Through systematic experiments on 19 video tasks involving 45 models across 15+ families, we observed pronounced variability in resilience. Notably, even MLLMs exceeding 70B parameters suffer severe performance drops, whereas select small models (\(<7\)B) demonstrate unexpectedly strong tampering detection, illustrating that size alone does not ensure robustness.

\paragraph{Impact on Research Community.} Beyond revealing these vulnerabilities, we believe that MVTamperBench offers actionable insights for refining model architectures and training pipelines, highlighting gaps that must be addressed before these systems can be reliably deployed in safety-critical settings. Our open-source framework will support reproducible evaluations and community-driven extensions, enabling researchers to integrate additional datasets or adapt tampering methods for domains like clickbait detection, content moderation, and surveillance feeds analysis.

\paragraph{Future Work.} Looking ahead, we plan to expand MVTamperBench with new tampering types (e.g., noise injection, frame shuffling) and domain-specific scenarios (e.g., healthcare, surveillance). By illuminating the nuanced impacts of video manipulations and guiding innovation in robust MLLM architectures, MVTamperBench lays a strong foundation for next-generation multimodal models capable of withstanding adversarial manipulations.

\section{Limitation}
\textbf{MVTamperBench} provides a robust framework for evaluating MLLM resilience against video tampering, but there are limitations that present opportunities for future work. 

First, while the benchmark evaluates five tampering techniques, expanding to additional manipulations/modalities to better capture subtle, and emerging techniques. Second,the dataset currently relies on limited datasets. Expanding the benchmark to incorporate videos from diverse sources, such as user-generated content, surveillance footage, or policy enforcement, would broaden its applicability and relevance to task-specific and domain-specific challenges. Third, while the current evaluation focuses on binary detection, future benchmarks could assess a model's ability to classify the specific tampering type, providing deeper insights into its robustness. Finally, scalability to closed-source and extremely large-scale models (\(>100\)B parameters) remains a challenge due to computational and cost constraints.

Addressing these limitations will enable \textbf{MVTamperBench} to further advance tampering detection and resilience in a broader range of applications.

\section*{Acknowledgement}
This work was partly supported by (1) the National Research Foundation of Korea(NRF) grant funded by the Korea government(MSIT)(RS-2024-00345398) and (2) the Institute of Information \& communications Technology Planning \& Evaluation(IITP) grant funded by the Korea government(MSIT)(RS-2020-II201373, Artificial Intelligence Graduate School Program (Hanyang University)).

\bibliography{acl}
\clearpage
\appendix

\section{Appendix}
\label{sec:appendix}

\subsection{MVTamperBench Details}
\label{sec:appendix_data}


\begin{table}[th!]
  \centering
  \begin{tabular}{p{3cm}p{4cm}}
    \hline
    \textbf{Dataset Name} & \textbf{Primary Scene Type and Unique Characteristics} \\
    \hline
    STAR\cite{Wu:24-star} & Indoor actions and object interactions \\
    \hline
    PAXION\cite{Wang:23-paxion} & Real-world scenes with nuanced actions \\
    \hline
    Moments in Time (MiT) V1\cite{Monfort:19-moments} & Indoor/outdoor scenes across varied contexts \\
    \hline
    FunQA\cite{Xie:25-funqa} & Humor-focused, creative, real-world events \\
    \hline
    CLEVRER\cite{Mao:22-clevrer} & Simulated scenes for object movement and reasoning \\
    \hline
    Perception Test\cite{Patraucean:24-perception} & First/third-person views for object tracking \\
    \hline
    Charades-STA\cite{charades2024} & Indoor human actions and interactions \\
    \hline
    MoVQA\cite{Zhang:23-movqa} & Diverse scenes for scene transition comprehension \\
    \hline
    VLN-CE\cite{Krantz:vlxce} & Indoor navigation from agent perspective \\
    \hline
    TVQA\cite{Lei:18-tvqa} & TV show scenes for episodic reasoning \\
    \hline
  \end{tabular}
  \caption{Summary of Datasets in MVTamperBench}
  \vspace{-1em}
  \label{tab:mvbench_datasets}
\end{table}

Table \ref{tab:mvbench_datasets} summarizes the datasets included in MVTamperBench, each contributing unique characteristics for robust tampering detection evaluation. We eliminate videos from the NTU dataset in MVTamperBench to avoid conflicting terms in the license. The dataset is then systematically expanded through the application of tampering effects to generate a comprehensive benchmark.

Table \ref{tab:benchmark_comparison} summarizes MVTamperBench's unique contributions compared to existing benchmarks \cite{hu2024visual,song2024moviechat,Li:24,Fu:24-VideoMME,xia2024cares,fei2024video} and use-cases which are spread across videos,images and documents \cite{agarwal2024techniques,agarwal2024domain,agarwal2024synthetic,agarwal2024enhancing,agarwal2024domain,agarwal-etal-2025-fs,patel2024llm,agarwal2023pseudo}. It also compares MVTamperBench with existing benchmarks, highlighting its focus areas, strengths, and unique contributions. We the growing interest area in MLLMS, new benchmarks that require special attention include Svbench \cite{yang2025svbench}, Video-MMLU \cite{song2025video}, SEA-VL \cite{cahyawijaya2025crowdsource}, MotionBench \cite{hong2025motionbench} and synthetic data generation techinques \cite{dua2024generating,dua2025generation,pabolu2024multi1,pabolu2024multi}.

\subsection{Ablation Studies}
We conduct ablation studies on key design decisions and prompt-engineering strategies to identify optimal configurations for constructing our benchmark. These experiments evaluate the effects of tampering duration, tampering position, and prompt formulation on model performance.

\subsubsection{Ablation: Tampering Position and Duration}
\label{sec:appendix-design}

To understand the influence of tampering characteristics on model performance, we conducted two ablation studies: (1) varying the duration of tampered segments, and (2) altering their position within the video timeline.


\paragraph{Tampering Position Ablation.} We varied the tampering position to occur after 25\%, 50\%, or 75\% of the video timeline (Table \ref{tab:ablation-location}). Detection performance remained largely stable across these conditions, indicating that most models are relatively insensitive to the specific location of tampering within the video. This suggests a consistent ability to maintain temporal coherence understanding regardless of when the manipulation occurs. We exclude tampering at the very beginning or end of videos, as these segments often coincide with natural scene transitions, delayed starts, or abrupt endings, which could introduce confounding noise into the benchmark.

\begin{table}[!th]
\centering
\small
\scalebox{0.95}{  
\begin{tabular}{llccc}
\hline
\textbf{Category} & \textbf{Model} & \textbf{25\%} & \textbf{50\%} & \textbf{75\%} \\
\hline
Low & \makecell[l]{Qwen2VL-7B} & 0.009 & 0.009 & 0.009 \\
    & \makecell[l]{LLaVaVideo-7B} & 0.006 & 0.006 & 0.006 \\
    & \makecell[l]{LLaVaOV-72B} & 0.0441 & 0.044 & 0.044 \\
\hline
Moderate & Aria & 0.7213 & 0.721 & 0.721 \\
         & \makecell[l]{Qwen2VL-72B} & 0.352 & 0.352 & 0.352 \\
         & \makecell[l]{Phi3.5Vision} & 0.707 & 0.707 & 0.707 \\
\hline
High & \makecell[l]{VILA1.5-40B} & 0.879 & 0.879 & 0.879 \\
     & \makecell[l]{InternVL2.5-8B} & 0.721 & 0.7210 & 0.721 \\
\hline
\end{tabular}
}
\caption{Impact of tampering position on F1 scores across models with varying robustness levels. Results are reported for tampering introduced at 25\%, 50\%, and 75\% positions within the video timeline.}
\label{tab:ablation-location}
\end{table}

\begin{table*}[!th]
\centering
\begin{adjustbox}{max width=\textwidth}
\begin{tabular}{m{0.15\textwidth}m{0.15\textwidth}m{0.2\textwidth}m{0.3\textwidth}m{0.3\textwidth}} 
\hline
\textbf{Benchmark} & \textbf{Scope (Image/Video)} & \textbf{Focus} & \textbf{Strengths} & \textbf{Unique Contributions} \\
\hline
BLINK & Image & Visual reasoning & Tests spatial relationships in images. & Introduces a framework for spatial reasoning with auxiliary sketches. \\
\hline
MUIRBENCH & Image & Multi-image understanding & Evaluates complex real-world scenarios. & Includes multi-image relations like narrative and complementary. \\
\hline
MVBench & Video & Temporal reasoning, event recognition & Assesses video understanding over time. & Focuses on video dynamics with temporal task coverage. \\
\hline
MMBench-Video & Video & Long-form video understanding & Handles multi-step event recognition in long videos. & Evaluates LVLMs on free-form QA using temporal reasoning. \\
\hline
Video-MME & Video & Multi-modal video understanding & Evaluates tasks like action recognition and captioning. & Combines multiple modalities for enhanced contextual understanding. \\
\hline
LongVU & Video & Spatiotemporal compression & Efficiently processes long videos with adaptive compression. & Novel spatiotemporal compression mechanism using cross-modal query. \\
\hline
MotionEpic & Video & Object tracking & Tracks object interactions across video frames. & Implements fine-grained spatial-temporal scene graph reasoning. \\
\hline
Wolf & Video & Video captioning & Improves video captioning with expert strategies. & Introduces CapScore for LLM-based caption evaluation. \\
\hline
Sharingan & Video & Action sequence extraction & Focuses on action recognition in desktop recordings. & Proposes differential and direct frame-based methods for user action extraction. \\
\hline
Video-of-Thought & Video & Step-by-step video reasoning & Excels in human-like video reasoning with chain-of-thought processes. & Integrates spatial-temporal scene graphs (STSG) for fine-grained reasoning. \\
\hline
CARES & Video & Scene comprehension and emotional analysis & Analyzes multi-modal emotional and contextual nuances. & Integrates context-aware emotional reasoning for enhanced video understanding. \\
\hline
Visual-Sketchpad & Image & Visual interaction and sketching tasks & Supports creative and interactive visual reasoning. & Bridges sketch-based reasoning with image analysis for enhanced user interaction. \\
\hline
MovieChat & Video & Conversational video understanding & Enhances video understanding with conversational context. & Introduces dialogue-based comprehension for temporal and narrative reasoning. \\
\hline
\textit{MVTamperBench \textbf{(Proposed)}} & Video & Tampered video detection & Robustly identifies tampered regions in video datasets. & Unique focus on domain-specific tampering scenarios with real-world applicability. \\
\hline
\end{tabular}
\end{adjustbox}
\caption{{Enhanced comparison between Video and Image Analysis Benchmarks, with unique contributions highlighted.}}
\label{tab:benchmark_comparison}
\end{table*}

\begin{table}[!th]
\centering
\small
\begin{tabular}{llccc}
\hline
\textbf{Category} & \textbf{Model} & \textbf{1s} & \textbf{2s} & \textbf{3s} \\
\hline
Low & Qwen2-VL-7B & 0.009 & 0.201 & 0.257 \\
    & LLaVa-Video-7B & 0.006 & 0.158 & 0.216 \\
    & LLaVa-\\
    & Onevision-72B & 0.044 & 0.257 & 0.249 \\
\hline
Moderate & Aria & 0.721 & 0.779 & 0.782 \\
         & Qwen2-VL-72B & 0.352 & 0.485 & 0.499 \\
         & Phi3.5-Vision & 0.707 & 0.763 & 0.769 \\
\hline
High & VILA1.5-40B & 0.879 & 0.898 & 0.901 \\
     & InternVL2.5-8B & 0.721 & 0.801 & 0.805 \\
\hline
\end{tabular}
\caption{Impact of tampering duration on F1 scores across models of varying robustness. Results are reported for tampering durations of 1s, 2s, and 3s.}
\label{tab:ablation-duration}
\vspace{-0.75em}
\end{table}

\paragraph{Tampering Duration Ablation.} We evaluated three tampering durations: 1s, 2s, and 3s. Results in Table \ref{tab:ablation-duration} indicate a consistent improvement in detection performance with increased tampering duration across all models. Lower-performing models (e.g., Qwen2-VL-7B, LLaVa-Video-7B) benefited more significantly, suggesting a dependency on longer anomalous intervals for effective detection. In contrast, top-performing models (e.g., VILA-1.5-40B, InternVL-2.5-8B) exhibited performance saturation, implying diminishing returns beyond a certain duration threshold.

\subsubsection{Ablation: Prompt Engineering}
\label{sec:appendix-prompt}

To determine the most effective prompt formulation for evaluating MLLM robustness in video tampering detection, we conducted a series of prompt-engineering experiments. Each variant prompts the model to assess whether the video has been tampered with, as aligned with the objective of our benchmark.

\begin{itemize}
    \item \textbf{Structured Prompt (used in benchmark)}: Explicitly lists tampering types.\\
    \textit{Prompt:} \textit{Does this video exhibit any signs of tampering, such as corruption, blackouts, rotated frames, repeated frames, or swapped frames?}

    \item \textbf{Generic Prompt}: Uses general phrasing that mimics natural user queries.\\
    \textit{Prompt:} \textit{Does this video exhibit any signs of tampering, manipulation, or inconsistency?}

    \item \textbf{Chain-of-Thought (CoT) Prompt}: Instructs the model to reason step by step before deciding.\\
    \textit{Prompt:} 
    \begin{quote}
    \textit{Scan the video step by step: 
    \begin{enumerate}
        \item Check each segment for visual glitches, repeated or missing content, or rotations.
        \item Check if any frame seems inconsistent with the rest of the video.
        \item Decide if any part looks manipulated or tampered.
    \end{enumerate}
    Does the video show any signs of tampering, manipulation, or inconsistency?}
    \end{quote}
\end{itemize}

\vspace{1em}

\begin{table}[h]
\centering
\small
\scalebox{0.95}{  
\begin{tabular}{llll}
\hline
\textbf{Category} & \textbf{Model} & \textbf{Prompt Type} & \textbf{F1 Score} \\
\hline
\multirow{9}{*}{Low} 
    & \multirow{3}{*}{Qwen2VL-7B} & Structured & 0.009 \\
    &                               & Generic    & 0.001 \\
    &                               & CoT        & 0.001 \\ \cline{2-4}
    & \multirow{3}{*}{LLaVaVideo-7B} & Structured & 0.006 \\
    &                               & Generic    & 0.002 \\
    &                               & CoT        & 0.001 \\ \cline{2-4}
    & \multirow{3}{*}{LLaVaOV-72B} & Structured & 0.044 \\
    &                               & Generic    & 0.001 \\
    &                               & CoT        & 0.001 \\
\hline
\multirow{9}{*}{Moderate}
    & \multirow{3}{*}{Aria} & Structured & 0.721 \\
    &                       & Generic    & 0.151 \\
    &                       & CoT        & 0.148 \\ \cline{2-4}
    & \multirow{3}{*}{Qwen2VL-72B} & Structured & 0.352 \\
    &                             & Generic    & 0.009 \\
    &                             & CoT        & 0.010 \\ \cline{2-4}
    & \multirow{3}{*}{Phi3.5Vision} & Structured & 0.707 \\
    &                              & Generic    & 0.201 \\
    &                              & CoT        & 0.208 \\
\hline
\multirow{6}{*}{High}
    & \multirow{3}{*}{VILA1.5-40B} & Structured & 0.879 \\
    &                              & Generic    & 0.458 \\
    &                              & CoT        & 0.466 \\ \cline{2-4}
    & \multirow{3}{*}{InternVL2.5-8B} & Structured & 0.721 \\
    &                                 & Generic    & 0.386 \\
    &                                 & CoT        & 0.389 \\
\hline
\end{tabular}
}
\caption{F1 scores of models evaluated with three prompt types: Structured, Generic, and Chain-of-Thought (CoT). Structured prompts consistently yield the highest performance across all robustness categories.}
\label{tab:prompt_ablation}
\end{table}

\noindent\textbf{Findings:} Structured prompts consistently outperformed both generic and CoT prompts across all model tiers (Table \ref{tab:prompt_ablation}). Low- and moderate-performing models exhibited significant performance drops with more open-ended prompts, likely due to limited temporal reasoning capabilities and lack of explicit training. Generic prompts particularly led to frequent false positives, as models misinterpreted benign variability as tampering. These findings support the use of structured prompts to ensure consistent and interpretable evaluation.

\begin{figure*}[th!]
    \centering
    \includegraphics[width=\textwidth]{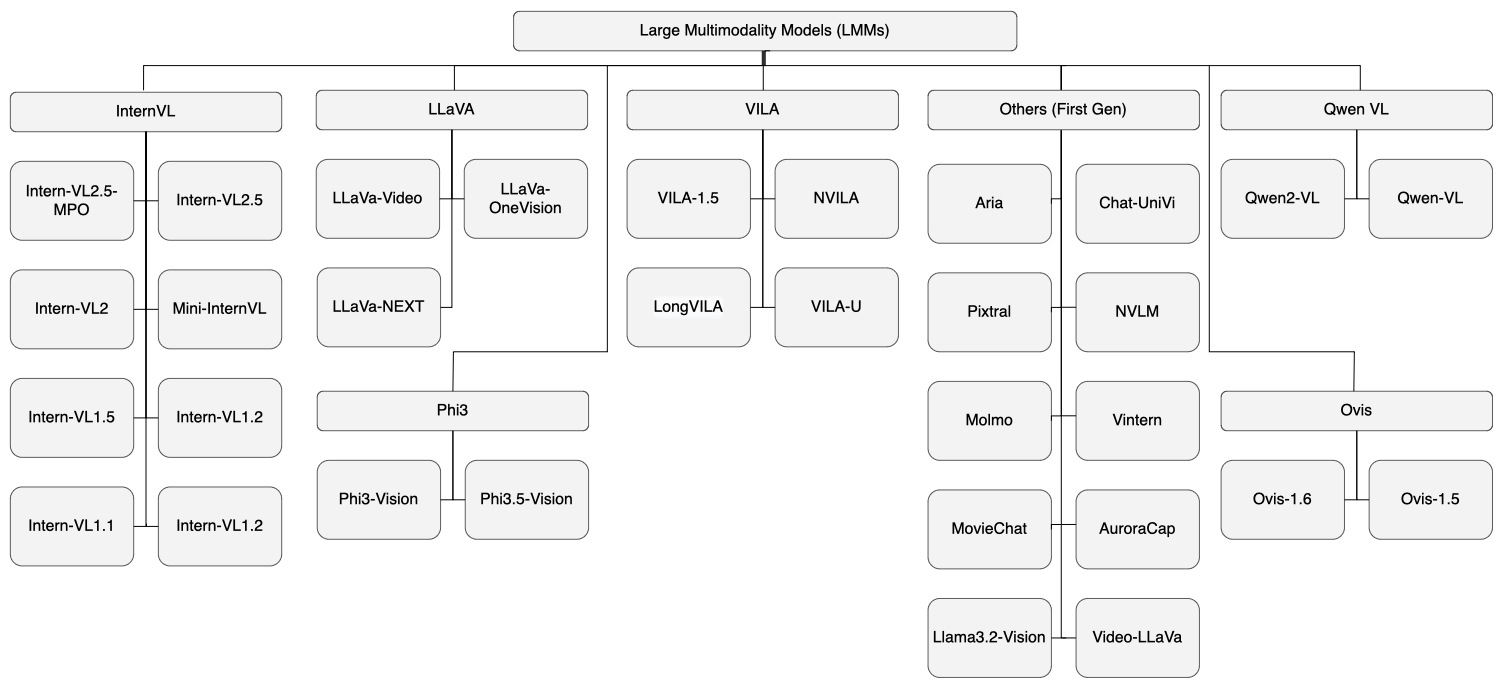}
    \caption{Taxonomy of Multimodal Large Language Models (MLLMs), organized by family, version, and first-generation releases.}
    \vspace{-1em}
    \label{fig:model_family}
\end{figure*}

\subsection{Overview of Multimodal Large Language Models (MLLMs)}
\label{sec:appendix_overview_lmm}

\paragraph{Model Families and Versions}
Figure \ref{fig:model_family} provides a comprehensive taxonomy of Multimodal Large Language Models (MLLMs), summarizing their diversity across families, versions, and first-generation releases. The diagram branches multiple versions of a model family (e.g., \textit{InternVL}, \textit{LLaVA}, \textit{VILA}, \textit{Phi3}, \textit{Ovis}) and separates earlier first-generation models (e.g., \textit{Video-LLaVA} \cite{lin2024videollavalearningunitedvisual,zhu2024languagebindextendingvideolanguagepretraining}, \textit{Vintern} \cite{doan2024vintern1befficientmultimodallarge}, \textit{LLama3.2-Vision} \cite{dubey2024llama,meta2024llama}) to reflect their evolutionary development, domain-specific popularity and capabilities in conversational systems \cite{pattnayak2025hybrid,pattnayak2025tokenizationmattersimprovingzeroshot,pattnayak9339review,patel2025sweeval}, finetuning \cite{thomas2025model}, and handling uses across documents \cite{yin2024continuous,agarwal2025techniques,panda2025out}, images and videos .

This taxonomy highlights the wide range of MLLMs currently available, organized as follows:
\begin{itemize}

    \item \textbf{InternVL Family}: Spanning models from \textit{Intern-VL1.1} \cite{chen2024far,gao2024mini,chen2024expanding} to the advanced \textit{Intern-VL2.5-MPO} \cite{chen2024internvl}, this family emphasizes efficiency and adaptability for diverse tampering and multimodal tasks. Successive iterations demonstrate marked improvements, particularly in handling temporal disruptions such as \textit{Dropping} and \textit{Substitution}. The \textit{InternVL2-5-8B} models showcases its ability to handle fine-grained spatiotemporal reasoning, highlighting its dominance in high-resource benchmarks.
    
    \item \textbf{LLaVA Family}: Starting with \textit{LLaVa-NEXT} \cite{li2024llava}, which struggled across most benchmarks, this family has evolved with models like \textit{LLaVa-OneVision} \cite{li2024llavaonevisioneasyvisualtask} and \textit{LLaVa-Video} \cite{zhang2024videoinstructiontuningsynthetic}, demonstrating significant improvements in task-specific video understanding through optimized pretraining and alignment techniques. Despite the advancements, \textit{LLaVa-OneVision \& LLaVa-Video} continues to face challenges in handling complex temporal disruptions, unlike \textit{Chat-UniVi}, which has emerged as a robust alternative.
    
    \item \textbf{VILA Family}: The \textit{VILA} series, including \textit{VILA-1.5} \cite{lin2024vila}, \textit{LongVILA}  \cite{xue2024longvila}, and \textit{VILA-U} \cite{wu2024vila}, demonstrates exceptional overall performance due to robust training pipelines and innovative architectures. Notably, \textit{VILA-40B} excels across benchmarks, which demonstrates advanced fine-grained and long-form video understanding capabilities, particularly in addressing tampering types like \textit{Masking} and \textit{Rotation}. Its architectural design allows it to efficiently process long-context visual inputs, setting a new standard for performance in large-scale benchmarks.
    
    \item \textbf{Phi3 Family}: Known for its scalable architecture and focus on real-world applicability, the \textit{Phi3-Vision} \cite{abdin2024phi} models struggle across tampering scenarios. The \textit{Phi3.5-Vision} model, in particular, highlights the benefits of improved tokenization and training strategies. It outperforms it's earlier versions by leveraging better alignment between visual and textual modalities.
    
    \item \textbf{Ovis Family}: Optimized for fine-grained visual-text alignment, \textit{Ovis} \cite{lu2024ovis} models leverage specialized datasets for tasks requiring high-resolution image interpretation and contextual reasoning. While the smaller version struggled with temporal coherence, subsequent larger model versions show promising improvements in spatial reasoning tasks like \textit{Masking} and \textit{Rotate}. Such enhancement results from better alignment between text and vision modalities.
    
    \item \textbf{First-Generation Releases}: Models such as \textit{Chat-UniVi} \cite{Jin_2024_CVPR}, \textit{Molmo} \cite{deitke2024molmopixmoopenweights}, \textit{NVLM} \cite{dai2024nvlm}, and \textit{Pixtral} \cite{agrawal2024pixtral12b} represent earlier efforts in video-language modeling. While some, like \textit{Molmo} and \textit{Aria} \cite{Li:24-aria}, continue to show competitive performance due to innovative training strategies, others, such as \textit{NVLM}, are limited by suboptimal optimization for temporal reasoning, which hinders their ability to adapt to tampering scenarios.
    
    \item \textbf{Qwen-VL Family}: The \textit{Qwen2-VL} \cite{wang2024qwen2} and \textit{Qwen-VL} \cite{yang2024qwen2} models are recent entrants that combine advanced architectures with scalable parameterization. They achieve strong results in grounding and visual reasoning tasks but struggle in detecting tampering across task and scenarios. Scaling the model size does help the performance but is still below the average performance of models in the study.
    
\end{itemize}

\begin{figure*}[ht!]
    \centering
    \begin{subfigure}{0.495\textwidth}
        \centering
        \includegraphics[width=\textwidth]{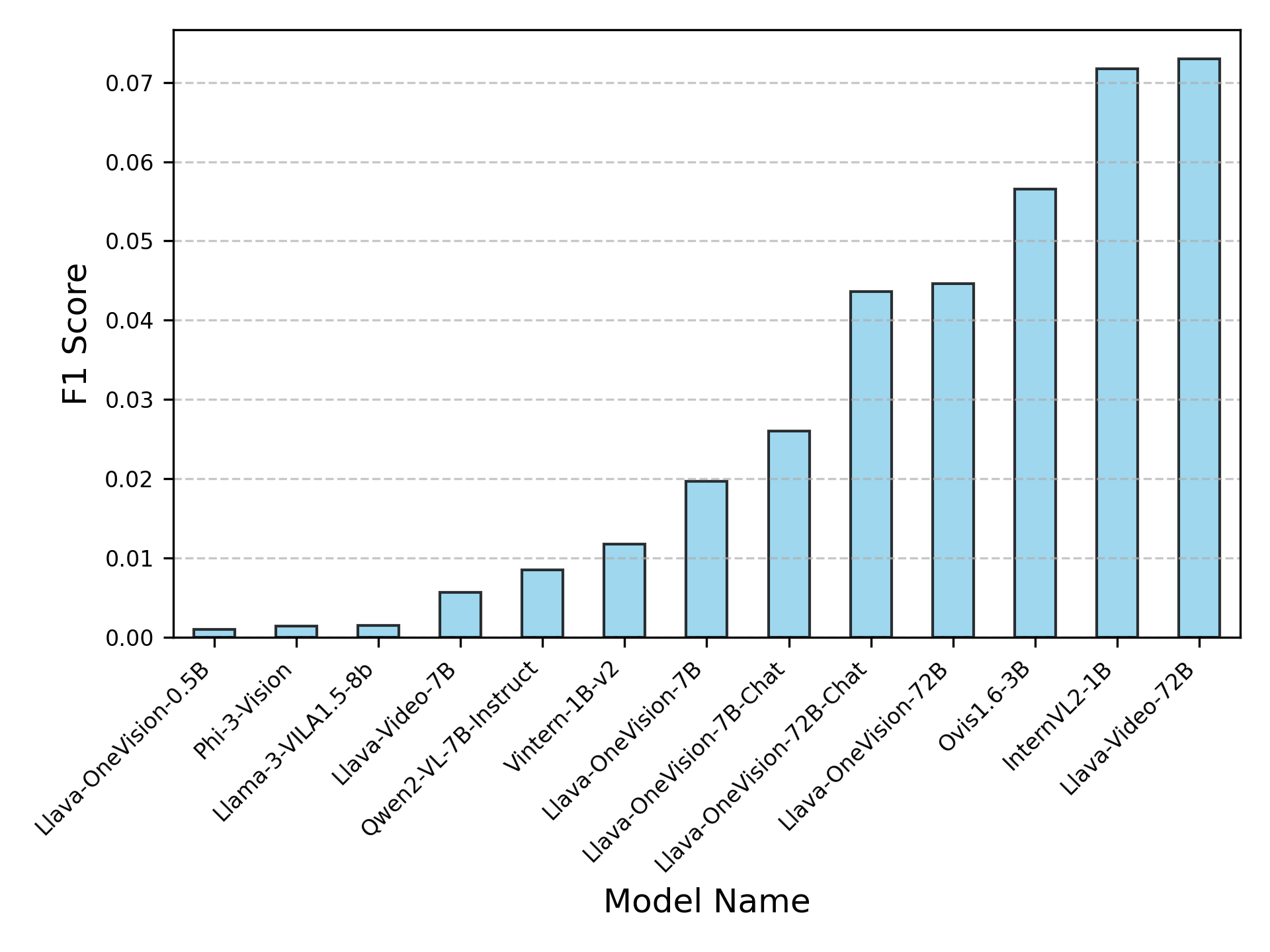}
        \caption{MLLMs in the Low Performing Category}
        \label{fig:graph_a}
    \end{subfigure}
    \hspace{-2em} 
    \begin{subfigure}{0.495\textwidth}
        \centering
        \includegraphics[width=\textwidth]{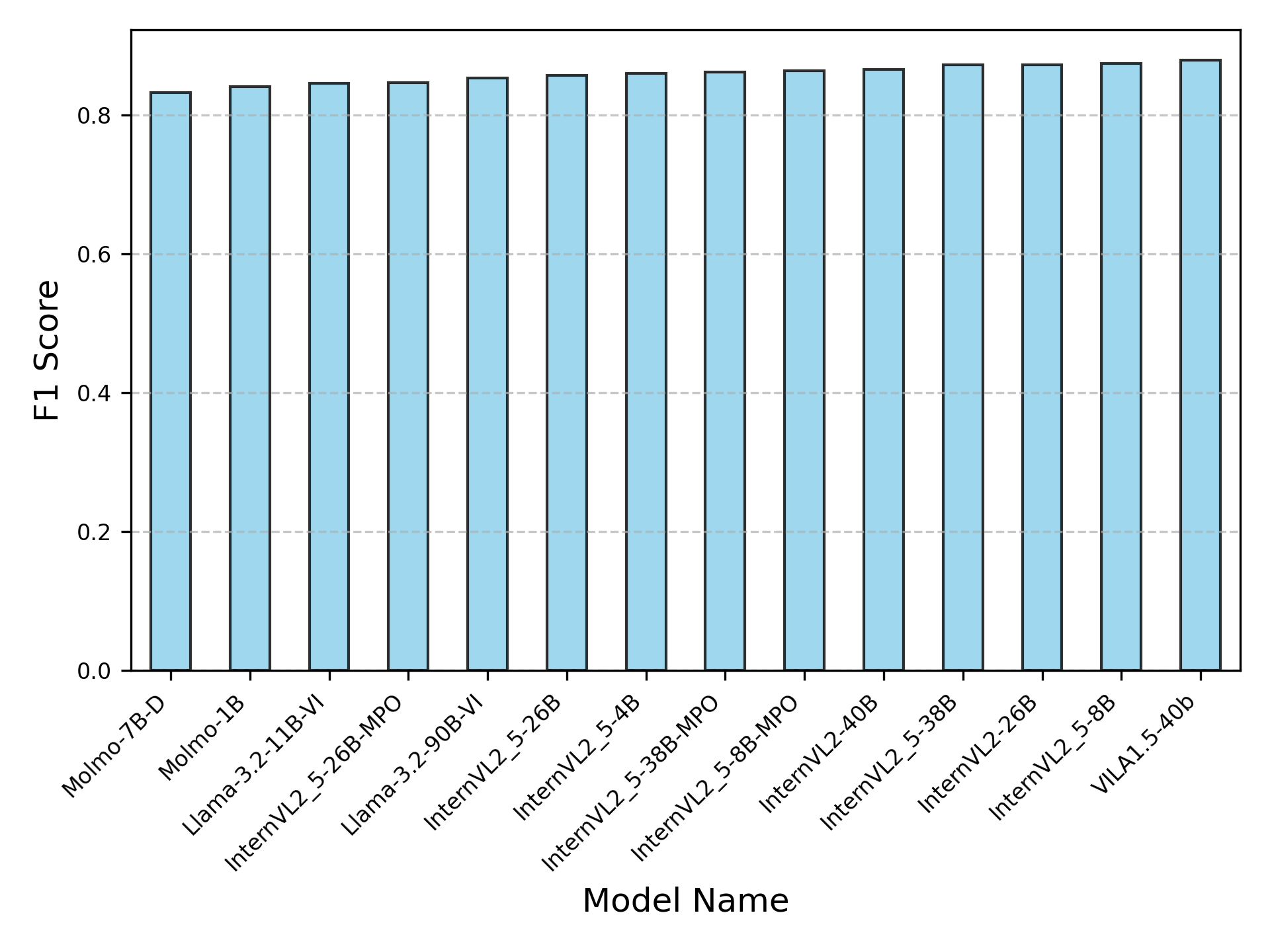}
        \caption{MLLMs in the High Performing Category}
        \label{fig:graph_b}
    \end{subfigure}

    \vspace{1em} 
    \begin{subfigure}{0.98\textwidth}
        \centering
        \includegraphics[width=\textwidth,height=0.8\textheight,keepaspectratio]{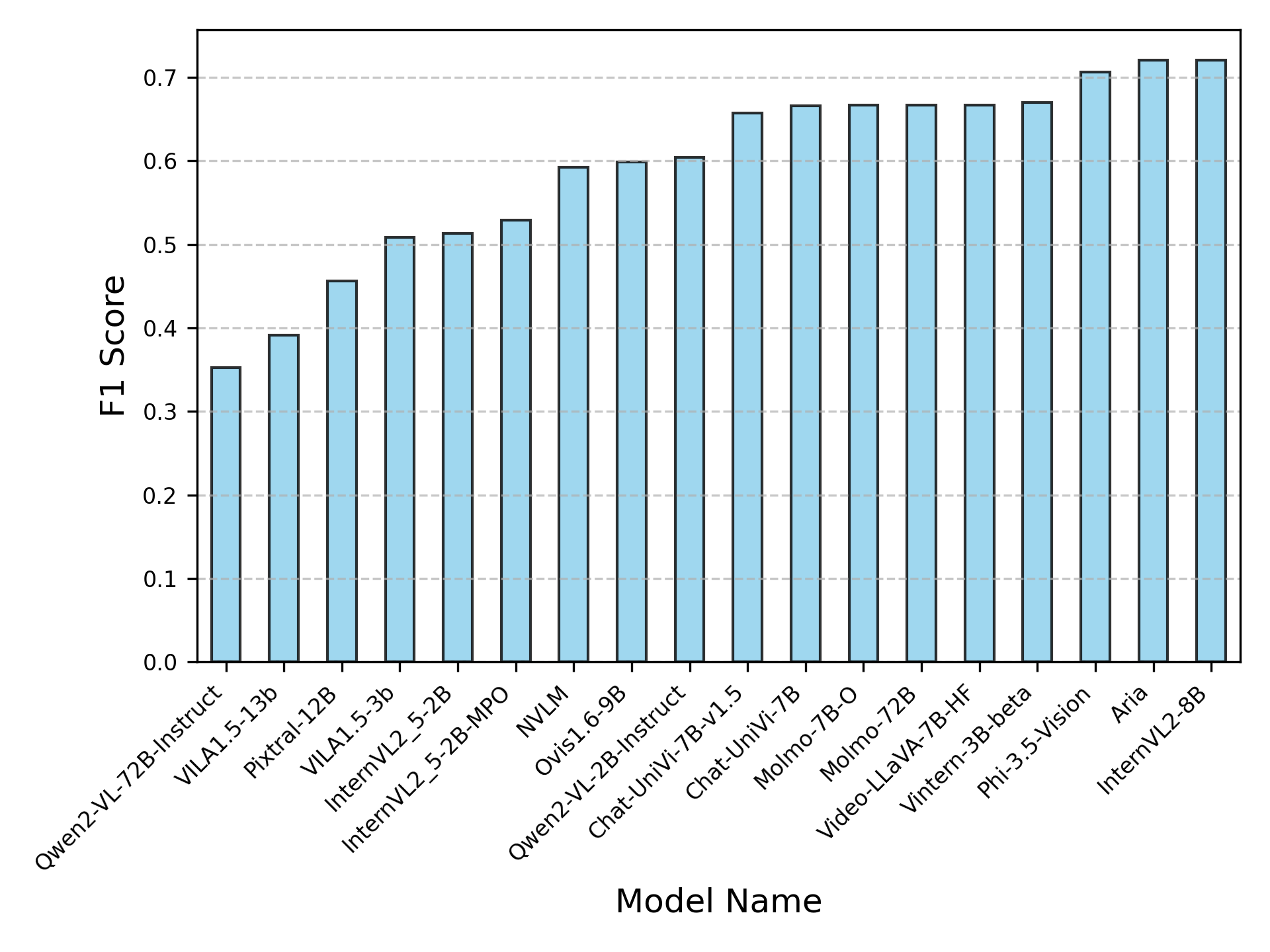}
        \caption{MLLMs in the Moderate Performing Category}
        \label{fig:graph_c}
    \end{subfigure}

    \caption{F1 (overall) performance of individual models across low, moderate, and high-performing categories. Models like \textit{InternVL-2.5} lead high-performing groups, while \textit{Llava-OneVision} models underperform consistently.}
    \label{fig:perf_model}
    \vspace{-1em}
\end{figure*}

\subsection{Extended Results \& Analysis}
\label{sec:appendix_overview_mllm}

\subsubsection{Analysis based on Performance Categories} 
\label{sec:appendix_model_perf}
Figure \ref{fig:perf_model} examines individual model performance across all categories, providing insights into trends among model families, versions, and architectures. 

For low-performing models (Figure \ref{fig:graph_a}), \textbf{Llava-OneVision} exhibits consistently weak performance across tampering types, even at larger parameter sizes (e.g., Llava-OneVision-72B). This suggests potential architectural and training data limitations, particularly for temporal coherence tasks. Interestingly, \textbf{Qwen2-VL-7B} underperforms significantly compared to its larger counterpart, \textbf{Qwen2-VL-72B}, which achieves a notable improvement. This indicates that increasing model size, combined with its training paradigm, positively impacts robustness for this family.

\begin{table*}[h!]
\centering
\small
\resizebox{\textwidth}{!}{%
\begin{tabular}{@{}lllp{2.8cm}p{3.5cm}p{4.3cm}@{}}
\toprule
\textbf{Perf.} & \textbf{Model} & \textbf{Vision} & \textbf{Language} & \textbf{Alignment Strategy} & \textbf{Training Strategy} \\
\midrule
High & Aria & SigLIP-400M & Aria-MoE & MoE decoder + lightweight ViT & 4-stage: language, multimodal, long-context, post-training \\
High & LLaMA & In-House & LLaMA-3.2 & — & Iterative SFT, rejection sampling, DPO on curated data \\
High & VILA & InternViT-6B & Yi-34B & Deep embedding projection + joint tuning & 3-stage: projector init, visual pretrain, instruction tuning \\
High & InternVL & InternViT-6B v2.5 & Qwen2.5 & QLLaMA middleware harmonization & Contrastive → generative → supervised tuning \\
\midrule
Moderate & Vintern & InternViT-300M & Qwen2.5 & MLP projector w/ visual instruction tuning & 2-stage: full-param + LoRA, cross-entropy loss \\
Moderate & Qwen2-VL & Custom ViT & Qwen2 & Dynamic resolution + M-RoPE & 3-stage: image–text pretrain + instruction tuning \\
Moderate & Pixtral & ViT-400M & Nemo-12B & Decoder fusion, ROPE-2D, sequence packing & Interleaved image–text pretraining \\
Moderate & NVLM & InternViT-6B & Qwen2 & Decoder/cross-attn/hybrid variants & Pre-align (frozen LLM), then multimodal SFT \\
Moderate & Molmo & CLIP ViT-L/14 & Qwen2-72B & Multi-crop connector + decoder LLM & Length-conditioned captions → multitask tuning \\
Moderate & Chat-UniVi & CLIP-ViT-L/14-336 & Vicuna-v1.5-7B & Unified visual tokens + multi-scale semantics & 2-stage: frozen pretrain + joint instruction tuning \\
\midrule
Low & PHI & CLIP ViT-L/14 & Phi-3 & Transformer decoder (block-sparse) & 2-phase: filtered web + synthetic \\
\bottomrule
\end{tabular}
}
\caption{Architecture, alignment, and training strategies for MLLMs stratified by robustness category.}
\label{tab:architecture-comparison}
\vspace{-1em}
\end{table*}

In the moderate-performing category (Figure \ref{fig:graph_c}), several trends emerge. \textbf{Llama3.2-11B} and \textbf{Llama3.2-90B} exhibit very similar performance despite the significant increase in size. These models, trained on the same recipe and dataset, highlight that merely increasing parameter count does not drastically enhance tampering detection capabilities. Another interesting observation is the improvement shown by \textbf{Phi3.5-Vision} and \textbf{Vintern-Beta} over their predecessors (\textbf{Phi3-Vision} and \textbf{Vintern}). This suggests that targeted architectural modifications or additional task-specific training significantly contribute to robustness. 

For high-performing models (Figure \ref{fig:graph_b}), \textbf{InternVL-2.5} dominates across tampering effects, with smaller versions (e.g., 4B) achieving comparable performance to their larger counterparts. This demonstrates that efficient architectural design and training strategies can offset limitations in model size. \textbf{VILA1.5-40B}, a model specifically designed for long-form video understanding, showcases exceptional robustness, emphasizing the importance of task-specific optimization. Notably, \textbf{Molmo-72B} and \textbf{NVLM-72B} exhibit below-average performance relative to other large models or their smaller counter parts, indicating inefficiencies in parameter utilization or potential overfitting to pretraining data and tasks.

\begin{figure*}[ht!]
    \centering
    \begin{subfigure}{0.495\textwidth}
        \centering
        \includegraphics[width=\textwidth]{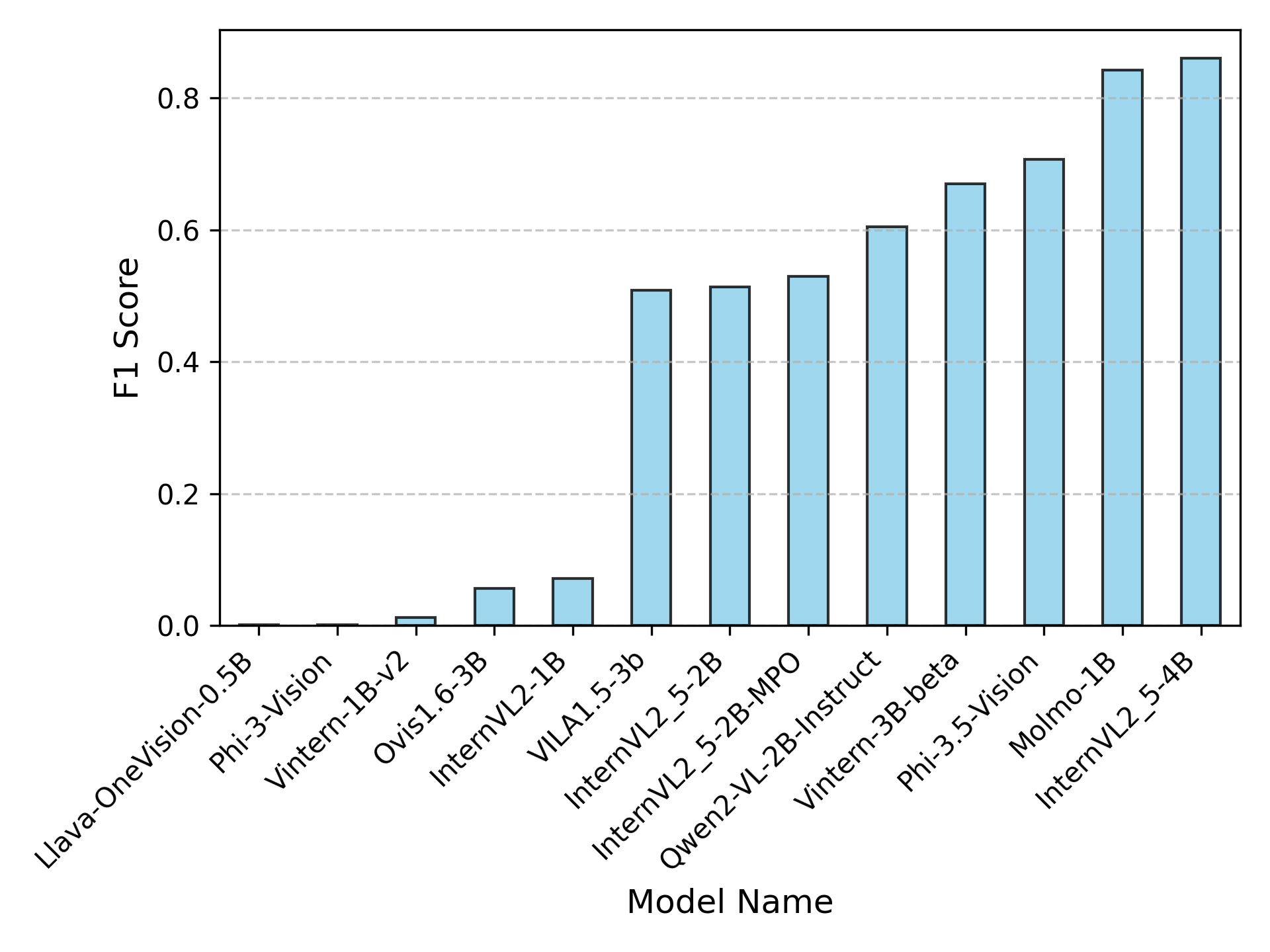}
        \caption{Performance Distribution of Small(<7B) MLLMs}
        \label{fig:graph_a_size}
    \end{subfigure}
    \hspace{-1em} 
    \begin{subfigure}{0.495\textwidth}
        \centering
        \includegraphics[width=\textwidth]{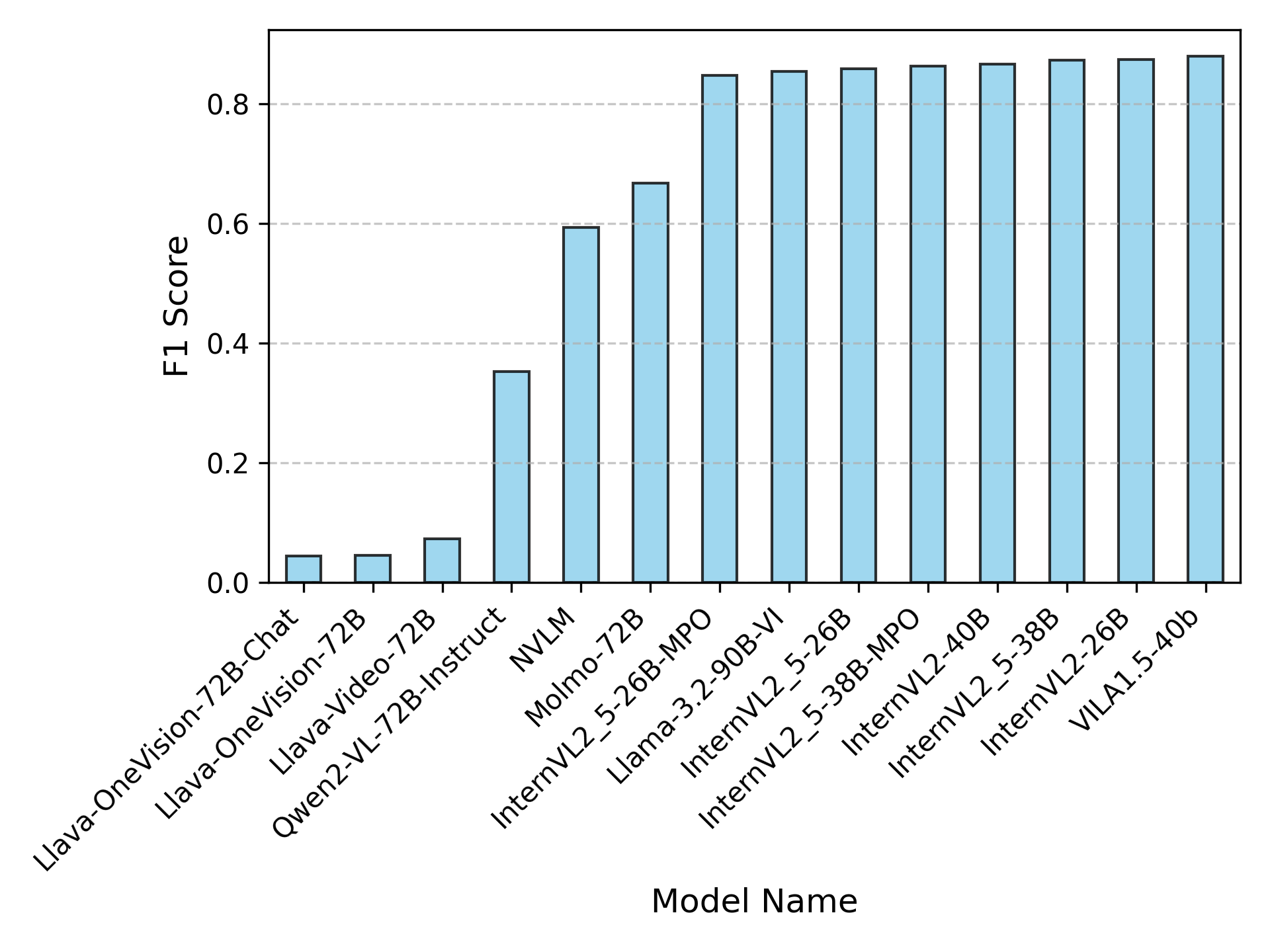}
        \caption{Performance Distribution of Large (>26B) MLLMs}
        \label{fig:graph_b_size}
    \end{subfigure}

    \vspace{1em} 
    \begin{subfigure}{0.98\textwidth}
        \centering
        \includegraphics[width=\textwidth,height=0.8\textheight,keepaspectratio]{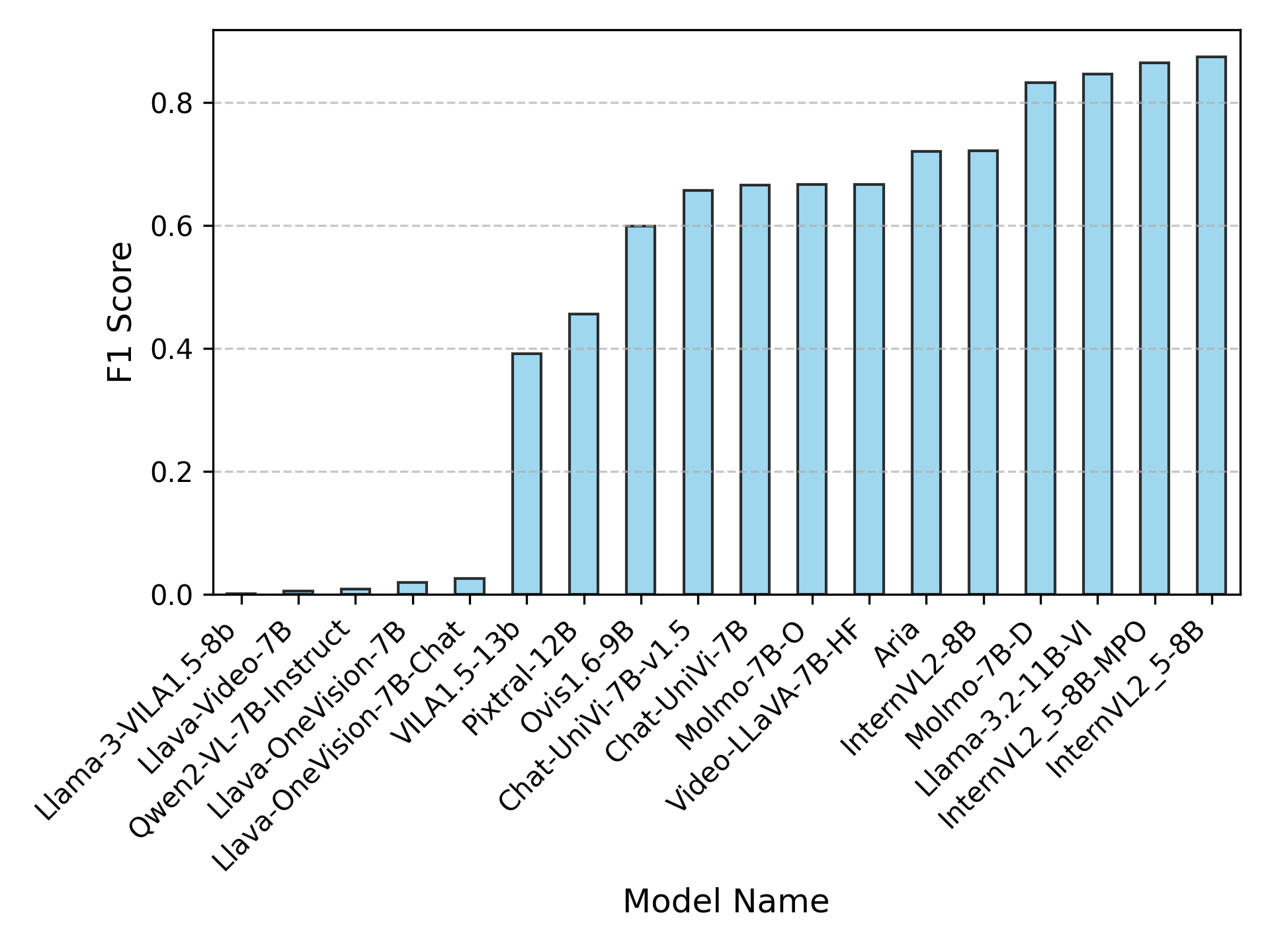}
        \caption{Performance Distribution of Medium(>=7B \& <=26B) MLLMs}
        \label{fig:graph_c_size}
    \end{subfigure}

    \caption{F1 (overall) performance of individual models across small, medium, and large  model size categories. Models like \textit{InternVL-2.5} lead high-performing groups, while \textit{Llava-OneVision} models underperform across categories.}
    \label{fig:size_perf_model}
\end{figure*}

Another noteworthy observation across categories is the consistency within certain model families. For example, the \textbf{Llava-Video} and \textbf{Chat-UniVi} models outperform their Llava-OneVision counterparts, demonstrating the importance of video-specific training for tampering detection. Conversely, while \textbf{Molmo-1B} excels among small models, its larger variant (\textbf{Molmo-72B}) does not scale proportionally in performance, reinforcing the need for efficient scaling and training strategies.

\subsubsection{Analysis based on Model Architectures}
\label{sec:appendix-architecture-trends}

To better understand model performance under video tampering, we conducted a comparative analysis of MLLMs' architectural design, alignment strategies, and training paradigms. Table~\ref{tab:architecture-comparison} summarizes key configurations across models stratified by their performance category (low, moderate, high).

Our findings suggest that robust MLLMs typically adopt deeper integration strategies between vision and language modalities. High-performing models like \textbf{VILA}, \textbf{InternVL}, and \textbf{Aria} utilize multi-stage training pipelines, explicit visual-text alignment layers (e.g., projectors or middleware), and instruction tuning on curated or human-annotated datasets. In contrast, moderate-performing models often rely on lightweight fusion (e.g., MLPs or prompt tuning) or lack post-alignment tuning stages. Low-performing models tend to use frozen CLIP encoders and shallow decoders with minimal multimodal alignment.

These insights support the hypothesis that robustness to spatiotemporal manipulations correlates with stronger cross-modal alignment and iterative supervision across training stages.

\subsubsection{Analysis based on Model Size}
\label{sec:appendix_model_size}

A closer examination of individual model trends across size categories (Figure \ref{fig:size_perf_model}) reveals several noteworthy patterns.

\paragraph{Small Models (<7B).}
Among small models (Figure \ref{fig:graph_a_size}), \textbf{Molmo-1B} demonstrates exceptional robustness, outperforming several medium-sized models and achieving consistency across all tampering types. Another notable small model, \textbf{Phi3.5-Vision}, shows drastic improvement over its predecessor, \textbf{Phi3-Vision}, highlighting the impact of architectural updates and extended task-specific training.

For the \textbf{VILA model family}, \textbf{VILA1.5-3B} performs better than many other small models, showcasing the benefits of targeted optimization for long-form video understanding. However, its performance lags behind \textbf{Molmo-1B} and \textbf{Phi3.5-Vision}, indicating room for improvement in handling complex tampering scenarios.

Interestingly, \textbf{InternVL-2.5-4B} matches or exceeds the performance of several medium-sized models, emphasizing the importance of efficient design over raw parameter counts. Models like \textbf{Llava-OneVision-7B}, however, remain among the weakest performers, suggesting that training approaches and architectural focus on static data limit their ability to handle tampering.

\paragraph{Medium Models (7B–26B).}
For medium-sized models (Figure \ref{fig:graph_c_size}), we observe considerable variability. Variants such as \textbf{Molmo-D} and \textbf{Molmo-O} exhibit similar performance, suggesting limited scalability within the family. Conversely, models like \textbf{InternVL-2.5-8B} maintain exceptional robustness, outperforming even larger models in the same category.

\textbf{Pixtral}, a medium-sized model that claims higher performance on other benchmarks, delivers average results here, highlighting that tampering-specific robustness requires distinct optimization strategies. Similarly, \textbf{Chat-UniVi} demonstrates an advantage over other Llava-family models, reinforcing the role of video-specific training in achieving higher tampering detection performance.

In the \textbf{VILA model family}, \textbf{VILA1.5-8B} underperforms compared to both smaller and larger VILA models, making it an interesting anomaly. This drop in performance could stem from suboptimal parameter tuning or an architectural bottleneck that affects scalability. Meanwhile, \textbf{VILA1.5-13B} and \textbf{VILA1.5-40B} continue to improve with increasing size, emphasizing the scalability of this family for long-form video tampering tasks.

\begin{figure*}[th!]
    \centering
    \includegraphics[width=\textwidth]{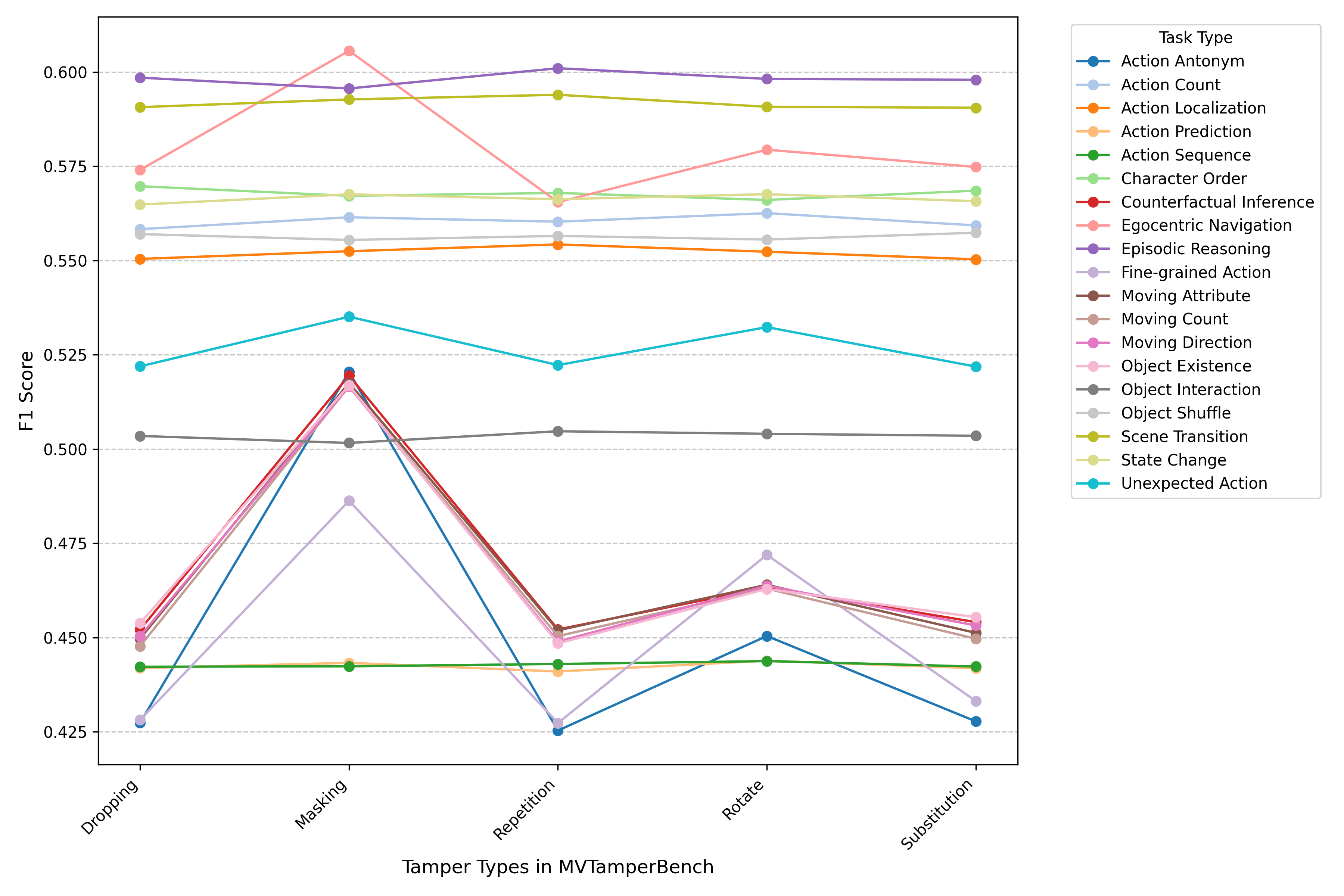}
    \caption{F1 (overall) scores across tampering types for task categories. \textit{Masking} is less disruptive, while \textit{Dropping} and \textit{Substitution} degrade performance in complex tasks like \textit{Counterfactual Inference}.}
    \vspace{-1em}
    \label{fig:tamper_f1_by_task_type}
\end{figure*}

\paragraph{Large Models (>26B).}

Among large models (Figure \ref{fig:graph_b_size}), we observe mixed performance trends.  While \textbf{InternVL-2.5} continues to dominate, its smaller variants (e.g., 4B, 8B) achieve comparable results, raising questions about the marginal benefits of scaling up within this family. \textbf{VILA1.5-40B}, specifically designed for long-form video understanding, ranks among the best-performing large models, showcasing the value of video-specific training \& optimization.

The \textbf{Qwen2-VL} family highlights the importance of scaling when paired with architectural optimization. \textbf{Qwen2-VL-72B} significantly outperforms its 7B counterpart, demonstrating the advantages of increased parameter counts and more extensive pretraining. However, its performance still lags behind other large models like \textbf{InternVL-2.5-40B} and \textbf{VILA1.5-40B}, suggesting potential inefficiencies in architecture or video-specific optimization.


The \textbf{Llama3.2-Vision} family offers a nuanced perspective on scaling. While \textbf{Llama3.2-11B} and \textbf{Llama3.2-90B} exhibit very similar performance, underscoring that increasing parameter count alone does not yield proportional performance gains. This trend reflects the importance of complementing scaling with architectural innovations and diverse, tamper-focused training data.

 While \textbf{InternVL-2.5} continues to dominate, its smaller variants (e.g., 4B, 8B) achieve comparable results, raising questions about the marginal benefits of scaling up within this family. Similarly, \textbf{Molmo-72B}, despite its size, fails to match the performance of its smaller counterparts, indicating inefficiencies in parameter utilization or overfitting to pretraining data. 
 
 In contrast, \textbf{Qwen2-VL-72B} delivers subpar results relative to its size but still outperforms its smaller sibling, \textbf{Qwen2-VL-7B}. This highlights that while scaling can improve performance, architectural and training advancements are critical for leveraging the full potential of larger models.

\subsubsection{Analysis across Video Task Types}
\label{sec:appendix_video_task}

\paragraph{Insights across Tampering Types.}
Figure \ref{fig:tamper_f1_by_task_type} reveals nuanced differences in performance across tampering types and video tasks. \textit{Masking} consistently emerges as the least disruptive effect, particularly in tasks like \textbf{Moving Direction} and \textbf{Object Existence}, where models rely more on contextual cues than on fine-grained spatial details. Conversely, \textit{Dropping} and \textit{Repetition} create the most significant challenges, particularly in tasks involving long-term temporal dependencies, such as \textbf{Action Sequence} and \textbf{Counterfactual Inference}.

Notably, the performance trends also vary across model categories. High-performing models demonstrate consistent robustness across all tasks and tampering types, while low-performing models exhibit the highest susceptibility to temporal and spatial disruptions. Moderate-performing models, on the other hand, display a mix of strengths and weaknesses, excelling in tasks with static or localized changes but struggling with tasks requiring temporal coherence.

\paragraph{Trends across Task Categories.}
Examining the task-wise trends, several key observations emerge:
\begin{itemize}
    \item Tasks involving \textbf{long-term temporal reasoning} (e.g., \textbf{Counterfactual Inference, Action Sequence}) are more sensitive to tampering, particularly \textit{Dropping}, \textit{Repetition}, and \textit{Substitution}, which disrupt narrative continuity.
    \item Tasks with \textbf{localized changes} (e.g., \textbf{State Change, Episodic Reasoning}) are less affected by tampering, as models rely on static visual cues.
    \item Tasks requiring \textbf{fine-grained spatial understanding} (e.g., \textbf{Fine-Grained Action}) show significant degradation under \textit{Rotation}, indicating a limitation in models’ ability to process spatial distortions.
    \item Performance disparities across tampering types highlight architectural strengths and weaknesses. For instance, \textbf{InternVL-2.5} excels in tasks like \textbf{Action Prediction} and \textbf{Counterfactual Inference}, owing to its advanced temporal reasoning capabilities.
\end{itemize}

\subsection{Benchmarking Efforts}

Our benchmarking efforts encompass 45 models across diverse categories and tampering scenarios, including Drop, Mask, Repeat, Rotate, and Substitute (Table \ref{tab:performance_metrics}). The results reveal significant variability in performance, influenced by model size, architecture, and training data.

Models such as \textit{VILA1.5-40B} and \textit{InternVL2.5-8B} emerged as top performers, achieving consistent resilience across all tampering types, with overall scores of 0.879 and 0.875, respectively. This highlights the importance of architectural innovations and advanced training techniques for tampering robustness. In contrast, early-generation models like \textit{LLaVA-OneVision} underperformed across all categories, with overall scores as low as 0.001, reflecting limitations in temporal coherence and token alignment.

Specialized models for video tasks, such as \textit{Chat-UniVi} and \textit{Video-LLaVA}, demonstrated substantial improvements over base \textit{LLaVA} models. \textit{Chat-UniVi-7B-v1.5} achieved an overall score of 0.658, significantly outperforming \textit{LLaVA-OneVision}, showcasing its ability to handle complex temporal manipulations. Meanwhile, \textit{Video-LLaVA-7B-HF} maintained robust performance across categories, further validating the effectiveness of unified tokenization and video-specific optimizations. However, \textit{LLaVA-Video}, despite efforts to improve alignment and pretraining, continues to struggle with certain tampering types, reflecting the challenges of adapting image-centric architectures to video modalities.

Interestingly, medium-sized models like \textit{Phi3.5-Vision} demonstrated notable performance improvements compared to earlier iterations such as \textit{Phi3-Vision}, indicating that scaling alone does not account for robustness gains. Specialized models like \textit{Ovis1.6-Gemma2-9B} showcased strengths in spatial tampering scenarios (e.g., Mask and Rotate) but struggled with temporal disruptions like Repeat and Substitute. This trend underscores the importance of task-specific optimizations.

\paragraph{Future Benchmarking Plans.}

While our analysis has covered an extensive set of models, several promising entries are yet to be evaluated. Models such as \textit{NVILA} \cite{liu2024nvila}, \textit{LongVILA} \cite{xue2024longvila}, and \textit{AuroraCap} \cite{chai2024auroracap} are currently being integrated into our benchmarking framework, with active collaborations underway to ensure seamless evaluations. Early insights suggest that these models could offer competitive performance in handling long-form video tampering and multimodal reasoning.

Additionally, we plan to expand the scope of evaluations for models like \textit{InternVL-1}, \textit{MovieChat} \cite{song2024moviechat,song2024moviechat+}, \textit{Vintern}, and future iterations of \textit{Chat-UniVi} and \textit{Video-LLaVA}. While \textit{Chat-UniVi} has shown impressive robustness across temporal and spatial tampering scenarios, and \textit{Video-LLaVA} continues to improve, exploring these models under additional tampering techniques will provide deeper insights into their limitations and areas for refinement.

The dynamic and evolving nature of our benchmarking framework ensures that future evaluations will continue to capture advancements in architectural design and tampering robustness.

\subsection{Key Findings}
\label{sec:appendix_key}

\paragraph{Consistent Performers.} The \textit{InternVL-2.5} series consistently outperforms other models by achieving strong F1 (overall) scores across all tampering types. Notably, even smaller variants like \textit{InternVL-2.5-4B} match the robustness of larger models. Such results highlight the efficiency of its architecture and training strategy.

\paragraph{Effect-specific Strengths.} Models such as \textit{Phi3.5-Vision} excel in detecting \textit{Masking} tampering, which indicates its specialized capabilities for handling visual obfuscations. Similarly, the \textit{VILA1.5-40B}, designed for long-form video understanding, excels in spatial-temporal tasks.

\paragraph{Weaker Models.} Certain models, including \textit{Llava-OneVision} variants, exhibit consistent weaknesses, particularly with temporal disruptions like \textit{Dropping} and \textit{Repetition}. This result may suggest limitations in their architectural designs and training paradigms.

\paragraph{Tampering Insights.} \textit{Dropping} and \textit{Repetition} emerge as the most challenging tampering types for all model categories, reflecting the difficulty of maintaining temporal coherence under such manipulations. In contrast, \textit{Masking} is relatively less disruptive, particularly for tasks relying on contextual cues.

\subsection{Discussion and Future Directions}
\label{sec:appendix_discussion}

The findings shed light on the importance of task-specific optimization and tamper-aware training for improving model robustness. Tasks that involve complex temporal dependencies or fine-grained spatial reasoning highlight critical gaps in current architectures, which need to better integrate temporal embeddings and multi-scale spatial attention mechanisms. MVTamperBench highlights the critical importance of robust architectures and diverse training data \& strategies for achieving tampering resilience. Below, we outline actionable insights and avenues for future exploration:

\paragraph{Expanding Benchmark Scope.} To enhance the benchmark's comprehensiveness, future iterations could evaluate additional model families, including emerging MLLMs. This will ensure a broader understanding of robustness trends.

\paragraph{Addressing Weak Models.} Models like \textit{Llava-OneVision} \& \textit{Qwen-2-VL} highlight the need for targeted improvements. Techniques such as adversarial training, task-specific fine-tuning, and architectural enhancements could improve performance.

\paragraph{Introducing New Tampering Types.} Expanding the benchmark to include tampering techniques such as localized masking, noise injection, and frame-level shuffling would provide a more nuanced evaluation of model resilience. Additionally, exploring domain-specific tampering types for critical applications like healthcare and surveillance could reveal context-dependent vulnerabilities.


\paragraph{Task-specific Insights.} Examining performance at a finer granularity—e.g., evaluating models on specific task categories or within specialized domains—could provide actionable guidance for training and optimization.

\paragraph{Scaling Considerations.} Observations such as diminishing returns for models like \textit{Molmo-72B} and \textit{Llama-3.2-90B-Vision} emphasize the need for efficient scaling strategies. Future work could explore methods for optimizing parameter utilization and balancing architectural complexity with training data diversity.

\paragraph{Integration with Real-world Applications.} Extending MVTamperBench to evaluate tampering resilience in real-world domains like media verification, misinformation detection, and legal forensics could uncover application-specific challenges and provide a pathway for practical deployments.

By exploring these directions, we hope that our MVTamperBench will evolve as a cornerstone benchmark for tampering detection, and further drive innovation in tamper-resilient MLLMs and foster trust in their real-world applications.

\begin{table*}[ht]
    \centering
    \small
    \scalebox{0.9}{\begin{tabular}{p{4cm}rcccccccc}
    \toprule
    \multirow{2}{*}{\textbf{Model}} & 
    \multirow{2}{*}{\textbf{Size}} & 
    \multirow{2}{*}{\textbf{Drop}} & 
    \multirow{2}{*}{\textbf{Mask}} & 
    \multirow{2}{*}{\textbf{Repeat}} & 
    \multirow{2}{*}{\textbf{Rotate}} & 
    \multirow{2}{*}{\textbf{Substitute}} & 
    \multirow{2}{*}{\textbf{Overall}} & 
    \textbf{Performance} & \textbf{Size} \\
    & & & & & & & & \textbf{Category} & \textbf{Category} \\
    \midrule
    Phi-3-Vision & 4 & 0.002 & 0.002 & 0.000 & 0.002 & 0.002 & 0.001 & Low & Small \\
    llava-onevision-qwen2-0.5b-ov  & 1 & 0.001 & 0.001 & 0.001 & 0.001 & 0.001 & 0.001 & Low & Small \\
    Llama-3-VILA1.5-8b  & 8 & 0.001 & 0.003 & 0.001 & 0.003 & 0.001 & 0.002 & Low & Medium \\
    llava\_video\_qwen2\_7b  & 7 & 0.006 & 0.006 & 0.005 & 0.005 & 0.006 & 0.006 & Low & Medium \\
    Qwen2-VL-7B-Instruct  & 7 & 0.005 & 0.004 & 0.004 & 0.025 & 0.005 & 0.009 & Low & Medium \\
    Vintern-1B-v2  & 1 & 0.013 & 0.011 & 0.012 & 0.011 & 0.011 & 0.012 & Low & Small \\
    llava-onevision-qwen2-7b-ov  & 7 & 0.016 & 0.038 & 0.010 & 0.019 & 0.016 & 0.020 & Low & Medium \\
    llava-onevision-qwen2-7b-ov-chat  & 7 & 0.020 & 0.050 & 0.013 & 0.027 & 0.020 & 0.026 & Low & Medium \\
    llava-onevision-qwen2-72b-ov-chat  & 72 & 0.021 & 0.125 & 0.018 & 0.034 & 0.021 & 0.044 & Low & Large \\
    llava-onevision-qwen2-72b-ov  & 72 & 0.020 & 0.132 & 0.017 & 0.033 & 0.021 & 0.045 & Low & Large \\
    Ovis1.6-Llama3.2-3B  & 3 & 0.057 & 0.064 & 0.050 & 0.053 & 0.058 & 0.057 & Low & Small \\
    InternVL2-1B  & 1 & 0.074 & 0.070 & 0.067 & 0.074 & 0.074 & 0.072 & Low & Small \\
    llava\_video\_qwen2\_72b  & 72 & 0.019 & 0.254 & 0.017 & 0.053 & 0.022 & 0.073 & Low & Large \\
    Qwen2-VL-72B-Instruct  & 72 & 0.279 & 0.544 & 0.282 & 0.374 & 0.284 & 0.352 & Moderate & Large \\
    VILA1.5-13b  & 13 & 0.378 & 0.422 & 0.383 & 0.391 & 0.382 & 0.391 & Moderate & Medium \\
    Pixtral-12B  & 12 & 0.452 & 0.469 & 0.453 & 0.454 & 0.453 & 0.456 & Moderate & Medium \\
    VILA1.5-3b  & 3 & 0.457 & 0.609 & 0.469 & 0.536 & 0.470 & 0.508 & Moderate & Small \\
    InternVL2\_5-2B & 
    2 & 0.500 & 0.538 & 0.505 & 0.523 & 0.499 & 0.513 & Moderate & Small \\
    InternVL2\_5-2B-MPO & 
    2 & 0.518 & 0.558 & 0.519 & 0.534 & 0.517 & 0.529 & Moderate & Small \\
    NVLM  & 72 & 0.588 & 0.595 & 0.598 & 0.595 & 0.588 & 0.593 & Moderate & Large \\
    Ovis1.6-Gemma2-9B  & 9 & 0.600 & 0.593 & 0.600 & 0.602 & 0.600 & 0.599 & Moderate & Medium \\
    Qwen2-VL-2B-Instruct  & 2 & 0.610 & 0.594 & 0.606 & 0.601 & 0.612 & 0.605 & Moderate & Small \\
    Chat-UniVi-7B-v1.5  & 7 & 0.662 & 0.642 & 0.663 & 0.661 & 0.661 & 0.658 & Moderate & Medium \\
    Chat-UniVi-7B  & 7 & 0.667 & 0.666 & 0.666 & 0.666 & 0.666 & 0.666 & Moderate & Medium \\
    molmo-7B-O-0924  & 7 & 0.667 & 0.667 & 0.667 & 0.667 & 0.667 & 0.667 & Moderate & Medium \\
    Video-LLaVA-7B-HF  & 7 & 0.667 & 0.667 & 0.667 & 0.667 & 0.667 & 0.667 & Moderate & Medium \\
    molmo-72B-0924  & 72 & 0.667 & 0.667 & 0.667 & 0.667 & 0.667 & 0.667 & Moderate & Large \\
    Vintern-3B-beta  & 3 & 0.670 & 0.669 & 0.674 & 0.669 & 0.670 & 0.670 & Moderate & Small \\
    Phi-3.5-Vision  & 4 & 0.676 & 0.822 & 0.677 & 0.682 & 0.677 & 0.707 & Moderate & Small \\
    InternVL2-8B  & 8 & 0.705 & 0.761 & 0.689 & 0.740 & 0.711 & 0.721 & Moderate & Medium \\
    Aria  & 25 & 0.717 & 0.738 & 0.716 & 0.719 & 0.716 & 0.721 & Moderate & Medium \\
    molmo-7B-D-0924  & 7 & 0.833 & 0.833 & 0.832 & 0.833 & 0.833 & 0.833 & High & Medium \\
    molmoE-1B-0924  & 1 & 0.842 & 0.842 & 0.842 & 0.842 & 0.842 & 0.842 & High & Small \\
    Llama-3.2-11B-Vision-Instruct  & 11 & 0.846 & 0.847 & 0.848 & 0.847 & 0.845 & 0.847 & High & Medium \\
    InternVL2\_5-26B-MPO & 
    26 & 0.848 & 0.848 & 0.847 & 0.848 & 0.848 & 0.848 & High & Large \\
    Llama-3.2-90B-Vision-Instruct  & 90 & 0.854 & 0.853 & 0.853 & 0.853 & 0.854 & 0.853 & High & Large \\
    InternVL2\_5-26B  & 26 & 0.858 & 0.858 & 0.858 & 0.858 & 0.858 & 0.858 & High & Large \\
    InternVL2\_5-4B  & 4 & 0.860 & 0.861 & 0.860 & 0.861 & 0.860 & 0.860 & High & Small \\
    InternVL2\_5-38B-MPO  & 38 & 0.857 & 0.868 & 0.859 & 0.868 & 0.859 & 0.862 & High & Large \\
    InternVL2\_5-8B-MPO  & 8 & 0.864 & 0.864 & 0.864 & 0.864 & 0.864 & 0.864 & High & Medium \\
    InternVL2-40B  & 40 & 0.863 & 0.870 & 0.864 & 0.870 & 0.865 & 0.866 & High & Large \\
    InternVL2\_5-38B  & 38 & 0.871 & 0.875 & 0.872 & 0.875 & 0.871 & 0.873 & High & Large \\
    InternVL2-26B  & 26 & 0.872 & 0.876 & 0.872 & 0.874 & 0.873 & 0.873 & High & Large \\
    InternVL2\_5-8B  & 8 & 0.875 & 0.875 & 0.875 & 0.875 & 0.875 & 0.875 & High & Medium \\
    VILA1.5-40b  & 40 & 0.879 & 0.880 & 0.878 & 0.880 & 0.879 & 0.879 & High & Large \\   
    \bottomrule
    \end{tabular}}
    \caption{Performance metrics for various models.}
    \label{tab:performance_metrics}
\end{table*}

\end{document}